\documentclass[runningheads]{llncs}
\usepackage{graphicx}

\usepackage{comment}
\usepackage{amsmath,amssymb} 
\usepackage{color}
\usepackage{epsfig}
\usepackage{graphicx}
\usepackage{mathrsfs}
\usepackage{url}
\usepackage{enumitem}
\usepackage{algpseudocode}
\usepackage{multirow}
\usepackage{booktabs}
\usepackage[pagebackref=false,breaklinks=true,letterpaper=true,colorlinks,urlcolor=blue,citecolor=blue,linkcolor=black,bookmarks=false]{hyperref}
\usepackage{subfigure}
\usepackage{setspace}
\usepackage{makecell}
\usepackage[linesnumbered,algoruled,boxed,lined]{algorithm2e}

\newcommand{\figdir}{figures}

\begin{document}
	\pagestyle{headings}
	\mainmatter
	\def\ECCVSubNumber{3150}  
	
	\title{Robust Tracking against Adversarial Attacks} 

	\titlerunning{Robust Tracking against Adversarial Attacks}
	%
	%
	\author{Shuai Jia\inst{1} \and
		Chao Ma\inst{1}\thanks{Corresponding author.}\and
		Yibing Song\inst{2}\and
		Xiaokang Yang\inst{1}}
	%
	 \authorrunning{S. Jia, C. Ma, Y. Song, and X. Yang}
	%
	
	\institute{MoE Key Lab of Artificial Intelligence, AI Institute, Shanghai Jiao Tong University\\
		\email{\{jiashuai,chaoma,xkyang\}@sjtu.edu.cn}
		\and Tencent AI Lab\\
		\email{yibingsong.cv@gmail.com}}
	
	
	\maketitle
	\begin{abstract}
		While deep convolutional neural networks (CNNs) are vulnerable to adversarial attacks, considerably few efforts have been paid to construct robust deep tracking algorithms against adversarial attacks. Current studies on adversarial attack and defense mainly reside in a single image. In this work, we first attempt to generate adversarial examples on top of video sequences to improve the tracking robustness against adversarial attacks. To this end, we take temporal motion into consideration when generating lightweight perturbations over the estimated tracking results frame-by-frame. 
		On one hand, we add the temporal perturbations into the original video sequences as adversarial examples to greatly degrade the tracking performance. On the other hand, we sequentially estimate the perturbations from input sequences and learn to eliminate their effect for performance restoration. We apply the proposed adversarial attack and defense approaches to state-of-the-art deep tracking algorithms. Extensive evaluations on the benchmark datasets demonstrate that our defense method not only eliminates the large performance drops caused by adversarial attacks, but also achieves additional performance gains when deep trackers are not under adversarial attacks. The source code is available at \url{https://github.com/joshuajss/RTAA}.
	\end{abstract}
	
	\section{Introduction}
	Recent years have witnessed the success of CNNs for numerous computer vision tasks. Along with the success, the problem of attacking CNN models using adversarial examples emerges recently. That is, small perturbations on input images can lead the pretrained CNN models to complete failures. A number of adversarial attack methods inject perturbations into input images to degrade the performance of CNNs on a wide range of vision tasks, such as image classification~\cite{Szegedy-iclr17-aa}, object detection~\cite{xie-iccv17-aa}, semantic segmentation~\cite{xiao-eccv18-aa}, and face recognition~\cite{dong-cvpr19-aa}. In view of the vulnerability of CNNs, the defense approaches~\cite{sun-cvpr19-aad,xie-cvpr19-aad} aim to improve the robustness of CNNs against adversarial attacks. 
    Despite the significant progress, current studies on adversarial attack and defense mainly rest in static images. Considerably less attention has been paid to generating adversarial examples on top of video sequences for robust deep tracking, where motion consistency between frames involves more challenges. 
	
	In this work, we start by investigating the vulnerability of the state-of-the-art deep trackers~\cite{nam-cvpr16-mdnet,li-cvpr18-siamrpn}, which pose object tracking as a sequential detection problem to distinguish between the target and background. The CNN classifiers are often updated online with positive and negative examples, which are collected according to the previously estimated tracking results. In our investigation, we do not modify existing deep trackers and keep their sampling schemes unchanged. For adversarial attack, we learn perturbations and inject them into input frames, yielding indistinguishable binary samples (i.e., a portion of the samples are incorrect). We use these binary adversarial examples to retrain CNN classifiers to degrade their performance. Specifically, we minimize the classification loss difference between the correct and incorrect binary samples.   
	When taking the temporal consistency between frames into consideration, we use the learned perturbations in the current frame to initialize the perturbation learning in the next frame. Applying the temporally generated perturbations for every frame further degrades the performance of deep trackers. 
	In addition to the CNN classifiers, existing deep trackers widely use a regression network to refine bounding boxes. We first attempt to randomly shift and rescale ground truth boxes to attack the regression network. Note that attacking the bounding box locations significantly differs from existing adversarial attack approaches on object detection~\cite{xie-iccv17-aa}, where perturbations are generated via considering mis-classifications.
	Fig.~\ref{fig:intro} shows such an example that the state-of-the-art deep trackers under adversarial attacks drift rapidly (see the first row).

	\renewcommand{\tabcolsep}{1pt}
	\def\swtwo{0.235\linewidth}
	\def\swone{0.60\linewidth}
	\def\swfour{0.35\linewidth}
	\begin{figure}[t]
		\begin{center}
			\includegraphics[width=\swone]{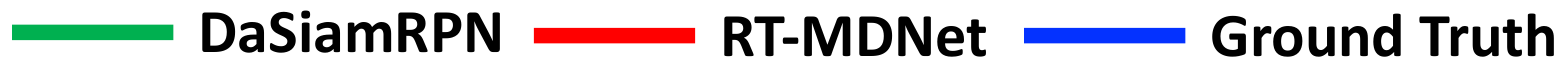}\\
				\begin{tabular}{ccccc}
					\rotatebox{90}{\bf \ \ \ \ Attack} &
					\includegraphics[width=\swtwo]{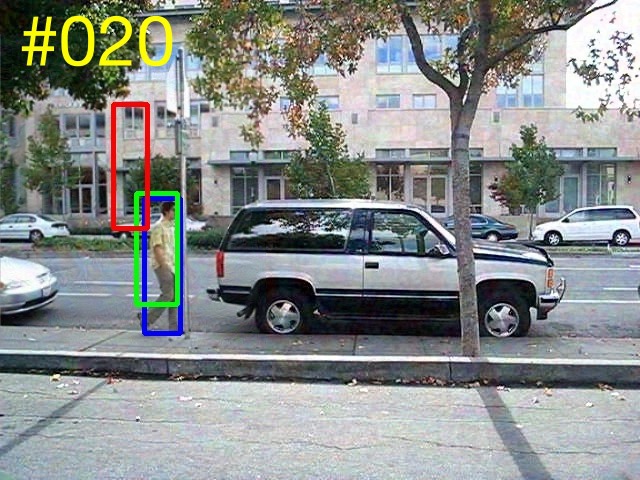}&
					\includegraphics[width=\swtwo]{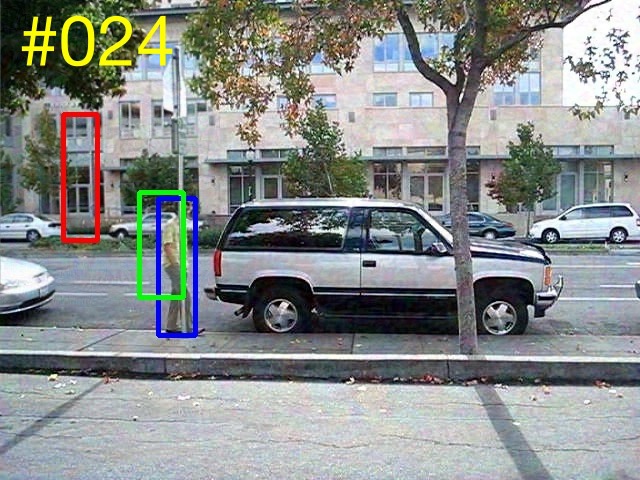}&
					\includegraphics[width=\swtwo]{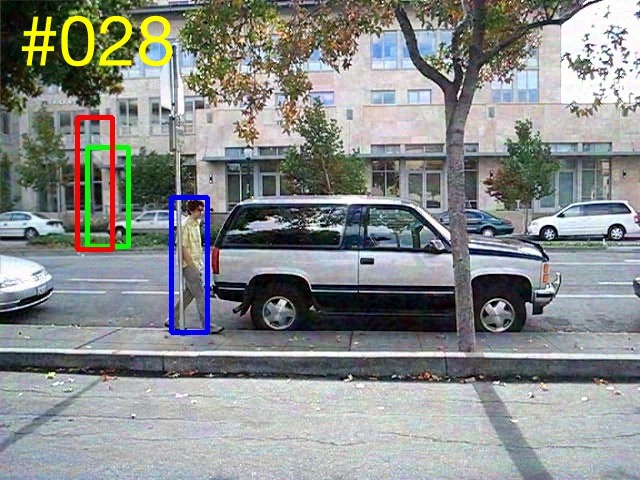}&
					\includegraphics[width=\swtwo]{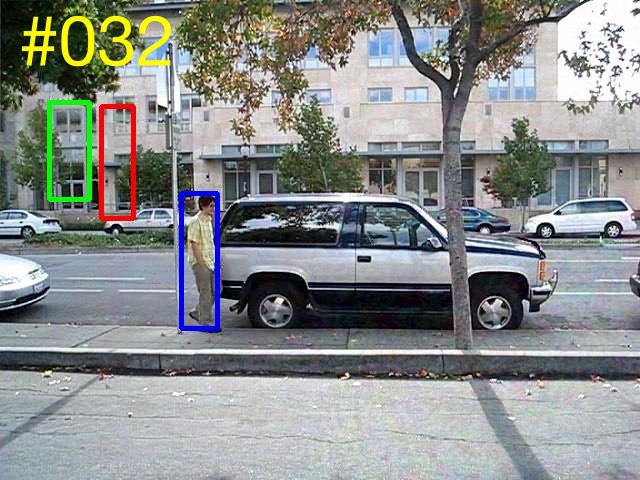}\\
					\rotatebox{90}{\bf \ \ \  Defense} &
					\includegraphics[width=\swtwo]{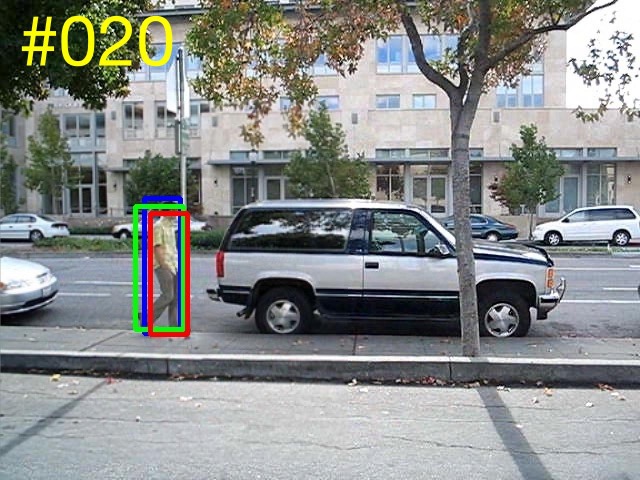}&
					\includegraphics[width=\swtwo]{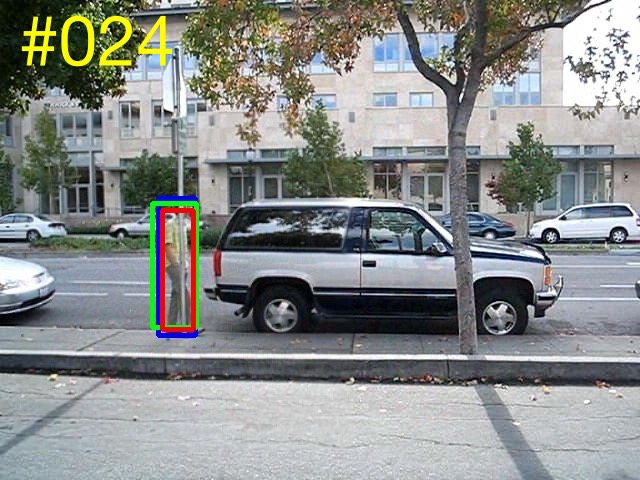}&
					\includegraphics[width=\swtwo]{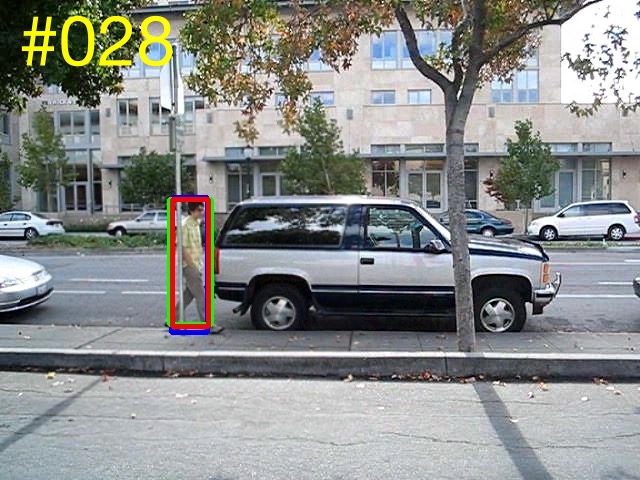}&
					\includegraphics[width=\swtwo]{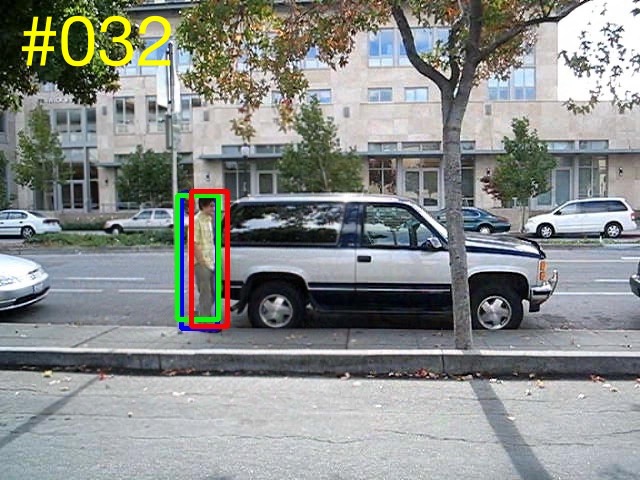}\\
				\end{tabular}
		\end{center}
		
		\caption{Adversarial attack and defense for visual tracking. 
			On top of the two state-of-the-art deep trackers DaSiamRPN~\cite{zhu-eccv18-dasiamrpn} and RT-MDNet~\cite{jung-eccv18-rtmdnet}, we learn to generate adversarial examples to attack and defend them on the \textit{David3} sequence~\cite{OTB-2015}.}
		\label{fig:intro}
	\end{figure}
	
	We step further to improve the robustness of deep trackers against adversarial attacks. Note that the adversarial perturbations are assumed to be unknown at this moment. Our motivation is to estimate the unknown perturbations in the input videos and learn to eliminate their effects during tracking. The estimation process is similar to the attack but the involved samples are different. As an example shown in Fig.~\ref{fig:intro}, we perform the proposed adversarial attack and defense approaches on two state-of-the-art deep tracking methods~\cite{zhu-eccv18-dasiamrpn,jung-eccv18-rtmdnet}. Extensive evaluations on large-scale benchmark datasets indicate the proposed defense approach is effective in improving the tracking robustness against adversarial attacks. When the trackers are not under adversarial attacks, the proposed defense scheme helps to achieve additional performance gains. This is because our defense is able to estimate the naturally existing adversarial perturbations during the image formation process.
	
	
	The main contributions of this work are summarized as follows:
	
	\begin{itemize}[noitemsep,nolistsep]
		\item We propose to generate adversarial examples to investigate the robustness of existing deep tracking algorithms. We inject dense adversarial perturbations into input video sequences in the spatiotemporal domain to degrade the tracking accuracy.
		\item We propose to defend deep trackers against adversarial attacks. We estimate the adversarial perturbations and eliminate their effect to alleviate performance drops caused by the adversarial attack. 
		\item We perform adversarial attack and defense on state-of-the-art deep trackers. The extensive evaluations on the benchmark datasets demonstrate the effectiveness of both attack and defense. Our defense can further advance the state-of-the-art deep trackers not under adversarial attacks.
	\end{itemize}

	\section{Related Works}
	\subsection{Deep Visual Tracking}
	Existing object tracking approaches can be generally categorized as one-stage regression based methods and two-stage detection based methods. Deep learning advances both categories of tracking methods significantly. The regression based methods typically learn correlation filters over CNN features to locate target objects as in~\cite{ma-iccv15-hcft}. Since that, numerous methods are proposed to improve tracking performance in different aspects, including feature hedging~\cite{qi-cvpr16-hdt}, continuous convolution~\cite{danelljan-eccv16-ccot}, particle filter integration~\cite{zhang-cvpr17-mcpf}, efficient convolution~\cite{danelljan-cvpr17-eco}, spatiotemporal regularization~\cite{li-cvpr18-learning}, and roi pooling~\cite{sun-cvpr19-roi}. Meanwhile, there are end-to-end learnable regression networks aiming to directly predict response maps~\cite{wang-iccv15-visual,held-eccv16-learning,valmadre-cvpr17-end,song-iccv17-crest,lu-18-deep,wang2019unsupervised,wang-udt-ijcv20} for localizing the target object.
	
	On the other hand, two-stage tracking-by-detection approaches first generate multiple proposals and then classify each as either the target or the background. The representative deep tracking-by-detection methods include multi-domain learning~\cite{nam-cvpr16-mdnet,jung-eccv18-rtmdnet}, ensemble learning~\cite{han-cvpr17-branchout}, adversarial learning~\cite{song-cvpr18-vital}, reciprocative learning~\cite{pu2018deep} and overlap maximization~\cite{danelljan-cvpr19-atom}. Recently, Siamese trackers~\cite{li-cvpr18-siamrpn,zhu-eccv18-dasiamrpn,li-cvpr19-siamrpn++,zhang-cvpr19-siamdw,bhat-iccv19-learning,ve-16-siamfc,lu-19-see} are prevalent due to their efficiency in online inference. The main difference between the Siamese trackers and other tracking-by-detection methods is that Siamese trackers typically do not online update CNN models while other methods do. In this work, we deploy the proposed adversarial attack and defense schemes on two representative state-of-the-art trackers including one Siamese  tracker~\cite{zhu-eccv18-dasiamrpn} without online update and one detector based tracker~\cite{nam-cvpr16-mdnet} with online update. Our goal is to illustrate the general effectiveness of adversarial attack and defense on deep trackers with or without online update.

	\subsection{Adversarial Attacks and Defense}
	Recent studies~\cite{goodfellow-iclr15-explaining,Szegedy-iclr17-aa} have shown that CNNs are vulnerable to adversarial examples. Despite the favorable performance on natural input images, the pretrained CNNs perform poorly given intentionally generated adversarial examples. Existing adversarial attack methods mainly fall into two categories: white-box and black-box attacks. The CNN models are assumed to be known in white-box attacks~\cite{goodfellow-iclr15-explaining,moosavi-cvpr16-deepfool,moosavi-cvpr17-universal}, whereas they are unknown in black-box attack~\cite{ilyas-arxiv18-black,dong-cvpr19-aa}. In addition to algorithmic attacks, physical attack methods generate real world objects to lead CNNs models to misclassifications. These are typically useful to examine the robustness of automotive driving in the road sign scenarios~\cite{kurakin-iclr17-adversarial,eykholt-cvpr18-robust,wiyatno-iccv19-physical}.
	
	Defending CNNs against adversarial attacks can be regarded as robustly learning CNNs with adversarial examples. 
	Attempts have been made to formulate defense as a denoising problem. From this perspective, the adversarial examples produce noise on CNN features to distract the network inference process. In~\cite{liao-cvpr18-ad,xie-cvpr19-aad}, denoising algorithms are proposed to eliminate the effect of noise. In addition, images are transformed to be non-differentiable in~\cite{guo-iclr18-ad} to resist adversarial attacks. Different from existing attack and defense methods, we attack both the classification and regression modules of deep trackers to decrease accuracy. Then, we gradually estimate adversarial perturbations and eliminate their effect on the input images without modifying existing deep trackers.
	
	\section{Proposed Algorithm}
	This section illustrates how to perform adversarial attack and defense for visual tracking. Given an input video sequence and a labeled bounding box in the initial frame, we generate adversarial examples spatiotemporally to decrease tracking accuracy. On the other hand, our defense learns to estimate unknown adversarial perturbations and eliminate their effect from input sequences. We deploy the proposed attack and defense algorithms on the tracking-by-detection framework. The details are presented in the following.
	
	\subsection{Adversarial Example Generation}\label{sec:aa}
	We generate adversarial perturbations based on the input frame and the output response of deep trackers, i.e., classification scores or regression maps. These perturbations are then added to the input frame for adversarial example generation. In the tracking-by-detection framework, deep trackers usually employ a CNN architecture containing two  branches. The sampled proposals are classified as either the target or background in the first branch, while the proposal axis is regressed in another branch for precise localization.
	A detailed illustration is referred to~\cite{li-cvpr18-siamrpn}. 
	We denote an input frame by $I$, the proposal number by $N$, the binary classification loss by $L_c$, the bounding box regression loss by $L_r$, the correct classification label and regression label by $p_c$ and $p_r$, respectively. Both labels $p_c$ and $p_r$ are predicted by the tracking results $S^{t-1}$ from the last frame, while $S^1$ is the ground-truth annotation in the initial frame. The original loss function of the tracking-by-detection network can be written as:
	\begin{equation}\label{eq:rpnLoss}
	\mathcal{L}(I,N,\theta)=\sum_{n=1}^{N}[L_c(I_n,p_c,\theta)+\lambda\cdot L_r(I_n,p_r,\theta)]
	\end{equation}
	where $I_n$ is one proposal in the image, $\lambda$ is a fixed weight parameter, and $\theta$ denotes the CNN parameters to be optimized during the training process.
		
	When generating adversarial perturbations, we expect CNNs to make inaccurate inference. We create a pseudo classification label $p^\star _c$ and a pseudo regression label $p^\star _r$. The adversarial loss is set to make $L_c$ and $L_r$ the same
	when we use correct and pseudo labels. The adversarial loss can be written as follows:
	\begin{equation}\label{eq:adverarialLoss}
	\begin{aligned}
	\mathcal{L}_{adv}(I,N,\theta) &= \sum_{n=1}^{N}\{[L_c(I_n,p_c,\theta)-L_c(I_n,p^\star _c,\theta)]\\
	&+ \lambda\cdot [L_r(I_n,p_r,\theta)-L_r(I_n,p^\star _r,\theta)]\}
	\end{aligned}
	\end{equation}
	where $\theta$ is fixed because the CNN is in the inference stage. The adversarial loss $\mathcal{L}_{adv}$ reflects the loss similarity between using correct and pseudo labels. When minimizing $\mathcal{L}_{adv}$, 
	the CNN predictions will be close to pseudo labels and the performance will degrade rapidly.
	
	We set pseudo labels specifically for each branch. In $p^\star _c$, there are two elements (i.e., 0 and 1) which indicate the probabilities of the input belonging to the target and background. We set $p^\star _c$ by reversing the elements of $p_c$ to confuse the classification branch. On the other hand, $p_r$ consists of four elements ($x_r$, $y_r$, $w_r$, $h_r$) representing the target location. We set $p^\star _r$ by adding a random distance offset and a random scale variation to $p_r$. Each element of $p^\star _r$ can be written as:
	\begin{eqnarray}\label{eq:axis}
	\nonumber  x^\star _r &=& x_r+\delta_{\rm offset}\\
	\nonumber  y^\star _r &=& y_r+\delta_{\rm offset}\\
	\nonumber  w^\star _r &=& w_r*\delta_{\rm scale}\\
	h^\star _r &=& h_r*\delta_{\rm scale}
	\end{eqnarray}
	where $\delta_{\rm offset}$ and $\delta_{\rm scale}$ indicate the random distance offset and random scale variation, respectively.

	\begin{algorithm}[t]
		\caption{Adversarial Example Generation}
		\KwIn{
		\ \ input video V with $T$ frames; target location $S^1$;}
		\KwOut{adversarial examples of $T$ frames; }
		
		\label{algo:1}
		\For{$t = 2 $  \KwTo $T$}
		{
			Get current frame $I^t_1$\;
			\If{$t \ne 2$}
			{Update $I^t_1$ via Eq.~\ref{eq:temporal}\;}
			
			\For{$m = 1$ \KwTo $M$}
			{
				\ Create $p_c$ and $p_r$ via $IoU$ ratios between proposals and target location $S^{t-1}$\;
				\  Create $p^\star _c$ by reversing elements of $p_c$\;
				\  Create $p^\star_r$ via Eq.~\ref{eq:axis}\;
				\  Generate adversarial loss via Eq.~\ref{eq:adverarialLoss}\;
				\  Update $I^t_m$ via Eq.~\ref{eq:final}\;
			}
			\Return $I^t_M$;
		}
	\end{algorithm}

	After computing the adversarial loss in Eq.~\ref{eq:adverarialLoss}, we take partial derivatives of the adversarial loss with respect to the input $I$. Formally, the partial derivative $r$ is computed as:
	\begin{equation}\label{eq:derivative}
	r=\frac{\partial \mathcal{L}_{adv}}{\partial I}.
	\end{equation}
	We pass $r$ into a $\operatorname{sign}$ function to reduce outlier effects. Given an input frame $I$, we take $M$ iterations to produce the final adversarial perturbations. The output of the last iteration is added into the input frame, which can be written as follows:
	\begin{equation}\label{eq:final}
	I_{m+1}=I_m+\alpha\cdot sign(r_m)
	\end{equation}
	where $\alpha=\frac{\epsilon}{M}$ is a constant weight, $\epsilon$ is the maximum value of the perturbations, $m$ indicates the iteration index, $I_m$ is the input frame for the $m$-th iteration, $\alpha\cdot sign(r_m)$ is the perturbations generated during the $m$-th iteration. The final adversarial example is $I_M$.

	As video frames are temporally coherent, we consider the adversarial attacks in the spatiotemporal domain.
	When there are $T$ frames in an input video sequence, we use the learned perturbations in the last frame as initialization for the current frame. Specifically, for the $t$-th frame, we use perturbations from the last frame to initialize $I^t$, which can be written as:
	\begin{equation}\label{eq:temporal}
	I_1^t=I_1^t+(I_{M}^{t-1}-I_{1}^{t-1})
	\end{equation}
	where $I_{M}^{t-1}-I_{1}^{t-1}$ is the perturbation from the last frame. Then we gradually update $I^t$ by using Eq.~\ref{eq:final} to generate the final perturbations for the $t$-th frame. The pseudo code is shown in Algorithm~\ref{algo:1}. Note that we use the $IoU$ metric~\cite{ren-pami16-faster} to assign proposal labels.

	\begin{algorithm}[t]
		\caption{Adversarial Example Defense}
		\KwIn{
			\ \ input video V with $T$ adversarial examples; target location $S^1$;}
		\KwOut{adversarial examples of $T$ frames; }
		
		\label{algo:2}
		\For{$t = 2 $  \KwTo $T$}
		{
			Get current frame $I^t_1$\;
			\If{$t \ne 2$}
			{Update $I^t_1$ via Eq.~\ref{eq:temporalD}\;}
			
			\For{$m = 1$ \KwTo $M$}
			{
				\  Create $p_c$ and $p_r$ via $IoU$ ratios between proposals and target location $S^{t-1}$\;
				\  Create $p^\star _c$ by reversing elements of $p_c$\;
				\  Create $p^\star_r$ via Eq.~\ref{eq:axis}\;
				\  Generate adversarial loss via Eq.~\ref{eq:adverarialLoss}\;
				\  Update $I^t_m$ via Eq.~\ref{eq:defense}\;
			}
			\Return $I^t_M$;
		}
	\end{algorithm}
		
	\subsection{Adversarial Defense}\label{sec:ad}
	We propose an adversarial defense method to recover tracking accuracy that is deteriorated via adversarial attacks. Our motivation comes from the adversarial perturbations which are accumulated during the iterations in each frame. From Eq.~\ref{eq:derivative}, we observe that perturbations originate from partial derivatives. Instead of adding perturbations to the input frame to decrease tracking accuracy, we estimate the perturbations and subtract them from the input frame gradually. As a result, the effect of perturbations will be eliminated to help CNNs to make the correct inference. We defend adversarial examples without updating CNNs.
	
		\begin{figure*}[t]
		\centering
		\includegraphics[width=1\linewidth]{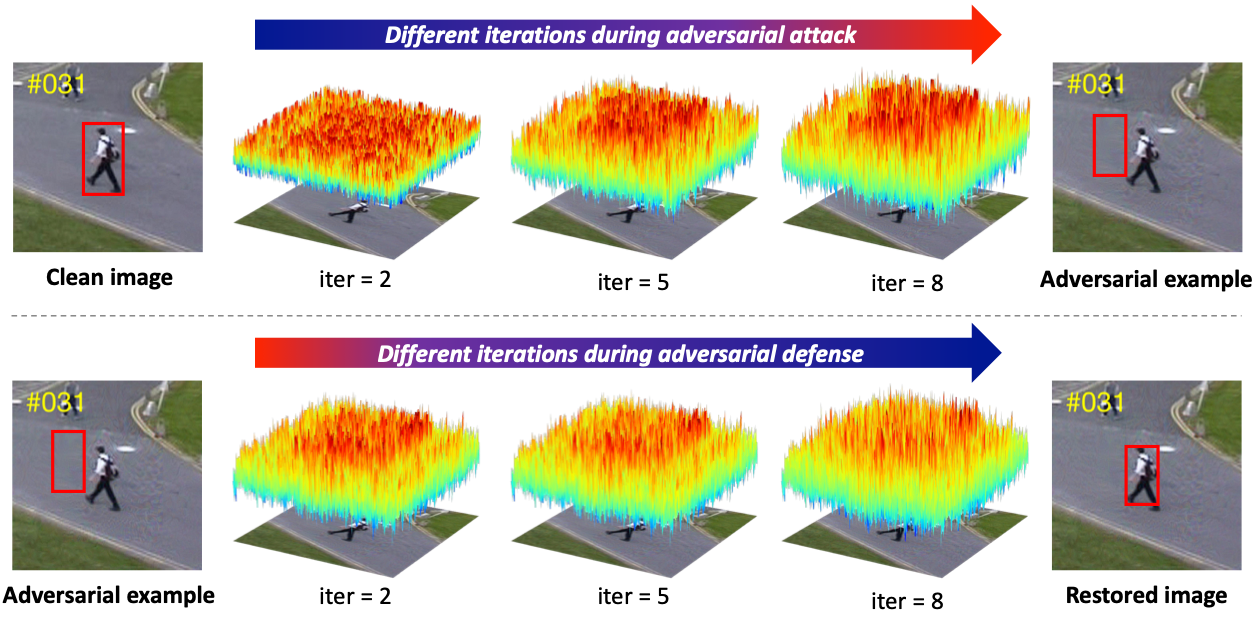}
		\caption{Variations of adversarial perturbations during attack and defense. The 3D response map above the image represents the difference between the clean image and the adversarial example at the current iteration. In adversarial attack, the perturbations increase along with training iterations. In adversarial defense, the perturbations decrease when training iteration increases. } \label{fig:method} 
	\end{figure*}

	Given an input frame $I$ with unknown adversarial perturbations, we first generate correct and pseudo labels according to the predicted location $S^{t-1}$ from the previous frame. The label generation process is similar to that in Sec.~\ref{sec:aa} except that the utilized proposals during defense are resampled. Then, we estimate the adversarial loss by using Eq.~\ref{eq:adverarialLoss} and compute the partial derivatives via Eq.~\ref{eq:derivative}. We apply partial derivatives $r$ on the input frame $I$ via the following operation:
	\begin{equation}\label{eq:defense}
	I_{m+1}=I_m-\operatorname*{Trunc}_{\beta\cdot r\in[-\hat{\alpha},\hat{\alpha}]}(\beta\cdot r)
	\end{equation}
	where $\beta$ is a constant weight, $\operatorname*{Trunc(\cdot)}$ is a truncation function to constrain the values of $\beta\cdot r$ within the range between $-\hat{\alpha}$ and $\hat{\alpha}$. The parameter $\hat{\alpha}$ resembles the parameter $\alpha$ in Eq.~\ref{eq:final}. Since the perturbation is unknown during defense, we empirically set different values for these two parameters. When the input videos contain $T$ frames, we still transfer the perturbations from the last frame to the current frame as initialization. For the $t$-th frame, we update it initially as:
	\begin{equation}\label{eq:temporalD}
	I_1^t=I_1^{t}-\gamma\cdot(I_{1}^{t-1}-I_{M}^{t-1})
	\end{equation}
	where $\gamma$ is a constant weight. The pseudo code of adversarial defense is shown in Algorithm~\ref{algo:2}.

	\subsection{Deployment of Deep Trackers}
	From the perspective of online update, existing tracking-by-detection methods involves two main stages. In the first stage, trackers do not update online while utilizing an offline pretrained CNN model. In the second stage, trackers collect samples online from previous frames to update the model. Note that trackers with online update tend to improve model adaption by collecting samples incrementally, which may help defend adversarial perturbations. We deploy the proposed adversarial attack and defense on two state-of-the-art trackers, DaSiamRPN~\cite{zhu-eccv18-dasiamrpn} and RT-MDNet~\cite{jung-eccv18-rtmdnet}. Details are presented in the following:

	{\flushleft\bf DaSiamRPN}. There are two output branches in DaSiamRPN to classify and regress proposals. During tracking, DaSiamRPN does not perform online update. When processing each frame, we follow Algorithm~\ref{algo:1} and Algorithm~\ref{algo:2} to generate and defend adversarial examples, respectively. As the inputs contain a template and search patch, we take partial derivatives with respect to the search patch when computing Eq.~\ref{eq:derivative} for both attack and defense.
	
	{\flushleft\bf RT-MDNet}. The CNN model of RT-MDNet only performs classification. During tracking, RT-MDNet online updates its model by collecting samples from previous frames. When processing each frame, we first generate adversarial examples and then perform prediction and model update. This configuration aims to analyze whether online update is effective to defend adversarial examples. We generate and defend adversarial examples using Algorithm~\ref{algo:1} and Algorithm~\ref{algo:2}, except that we remove the regression terms when computing Eq.~\ref{eq:adverarialLoss}. 
	
	{\flushleft\bf Visualizations}. We show how adversarial perturbations vary during different training iterations in Fig.~\ref{fig:method}. DaSiamRPN is our baseline tracker to be attacked and defended. Given an input frame, we visualize the perturbations during adversarial attack. Along with the training iterations, the variation of perturbations increases as well. The adversarial examples lead DaSiamRPN to drift rapidly. When defending this adversarial example, we observe that the variation of the perturbations decreases when training iteration increases. This indicates that our defense method is effective to estimate and exclude the adversarial perturbations, which helps to alleviate performance drops caused by adversarial attack.
	
	\section{Experiments}
	
	In this section, we evaluate the proposed attack and defense methods on benchmark datasets. We develop our methods on top of DaSiamRPN and RT-MDNet for all the experiments. The maximum variation value of each pixel in the perturbations is set to 10 (i.e., $\epsilon = 10$) for attack and set to 5 for (i.e., $\epsilon = 5$) defense, respectively. When computing $IoU$ ratios between proposals and the target location $S_{t-1}$, we follow the threshold setting of DaSiamRPN and RT-MDNet to draw training samples. The parameters of deep trackers are kept fixed during both adversarial attack and defense.
	
	\begin{figure}[t]
		\begin{center}
			\begin{tabular}{cc}
				\includegraphics[width=\swfour]{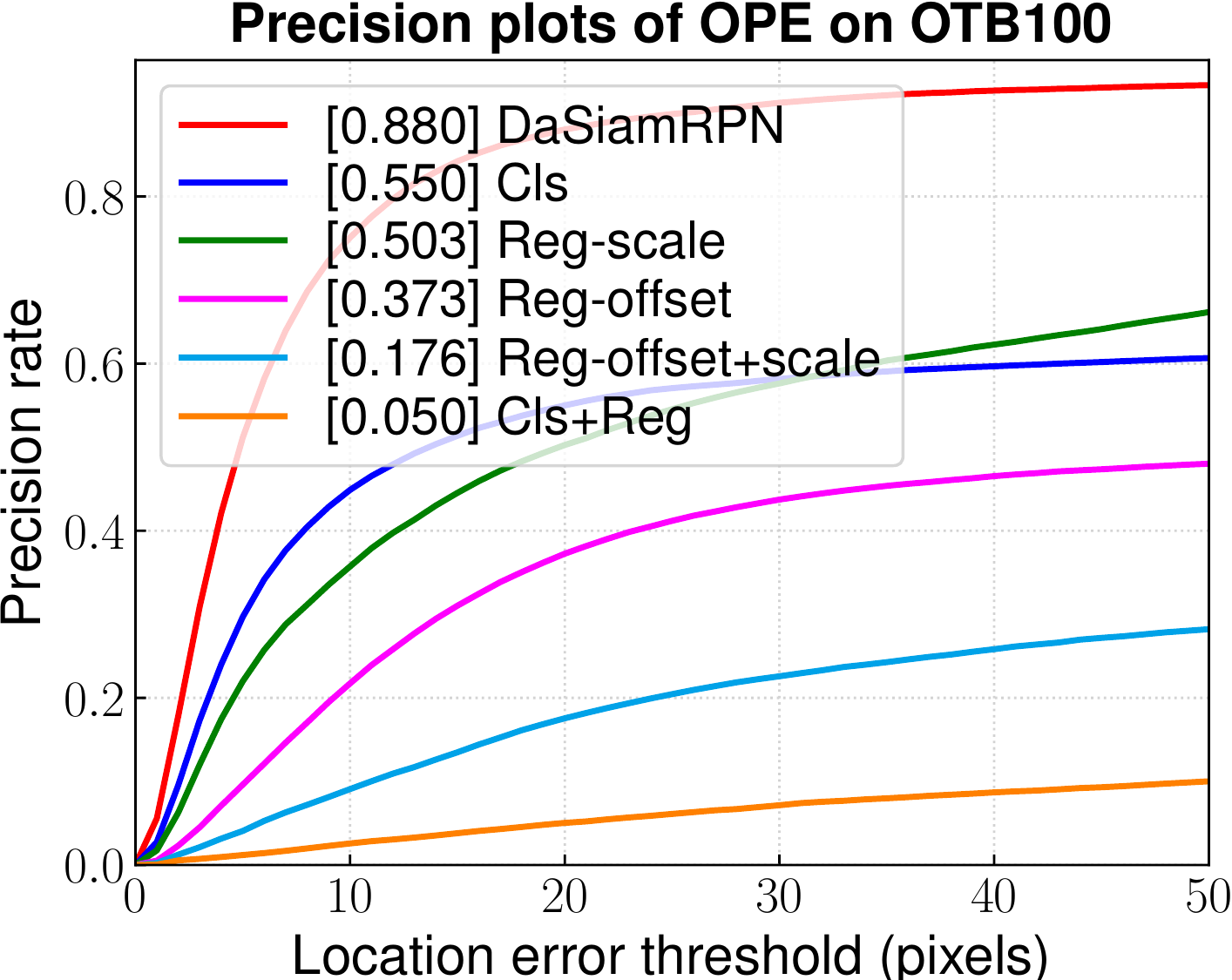}\ \ \ 
				\includegraphics[width=\swfour]{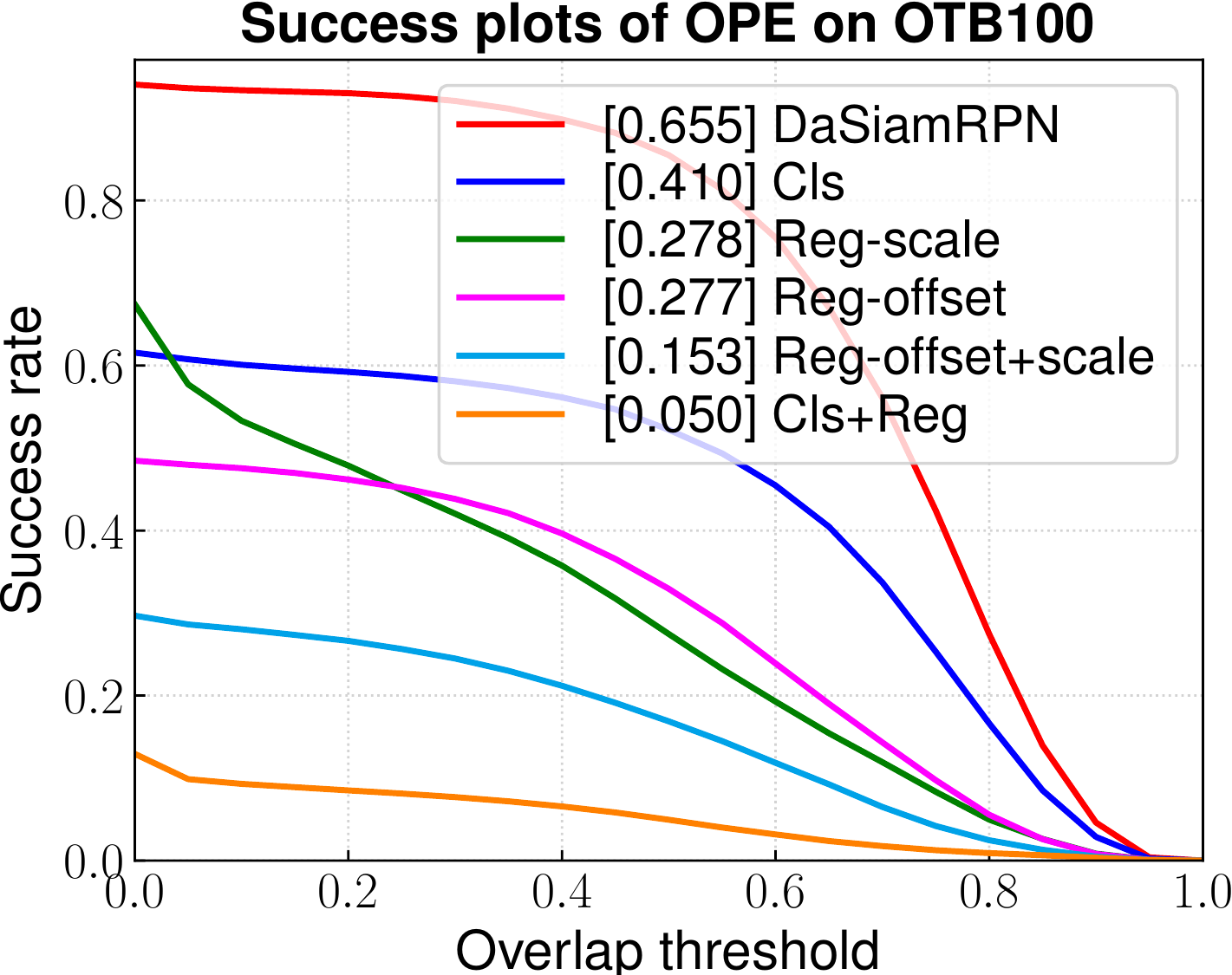}	
			\end{tabular}
		\end{center}
		\caption{Ablation studies of DaSiamRPN on the OTB100 dataset \cite{OTB-2015}. We denote Cls as the attack on the classification branch, Reg as the attack on the regression branch where there are offset and scale attacks.}
		\label{fig:ab1}
	\end{figure}

	\begin{figure}[t]
	\begin{center}
		\begin{tabular}{cc}
			\includegraphics[width=\swfour]{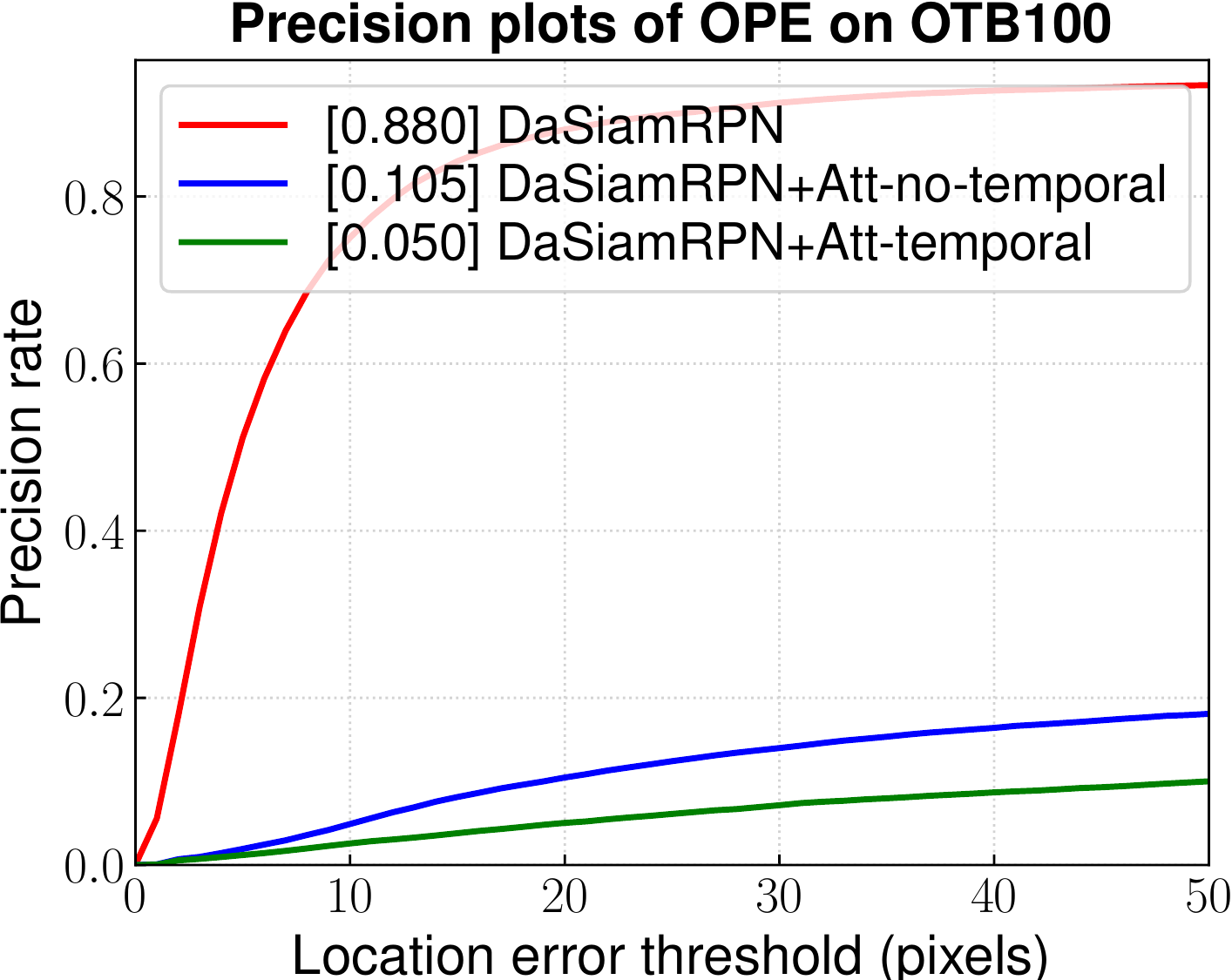}\ \ \ 
			\includegraphics[width=\swfour]{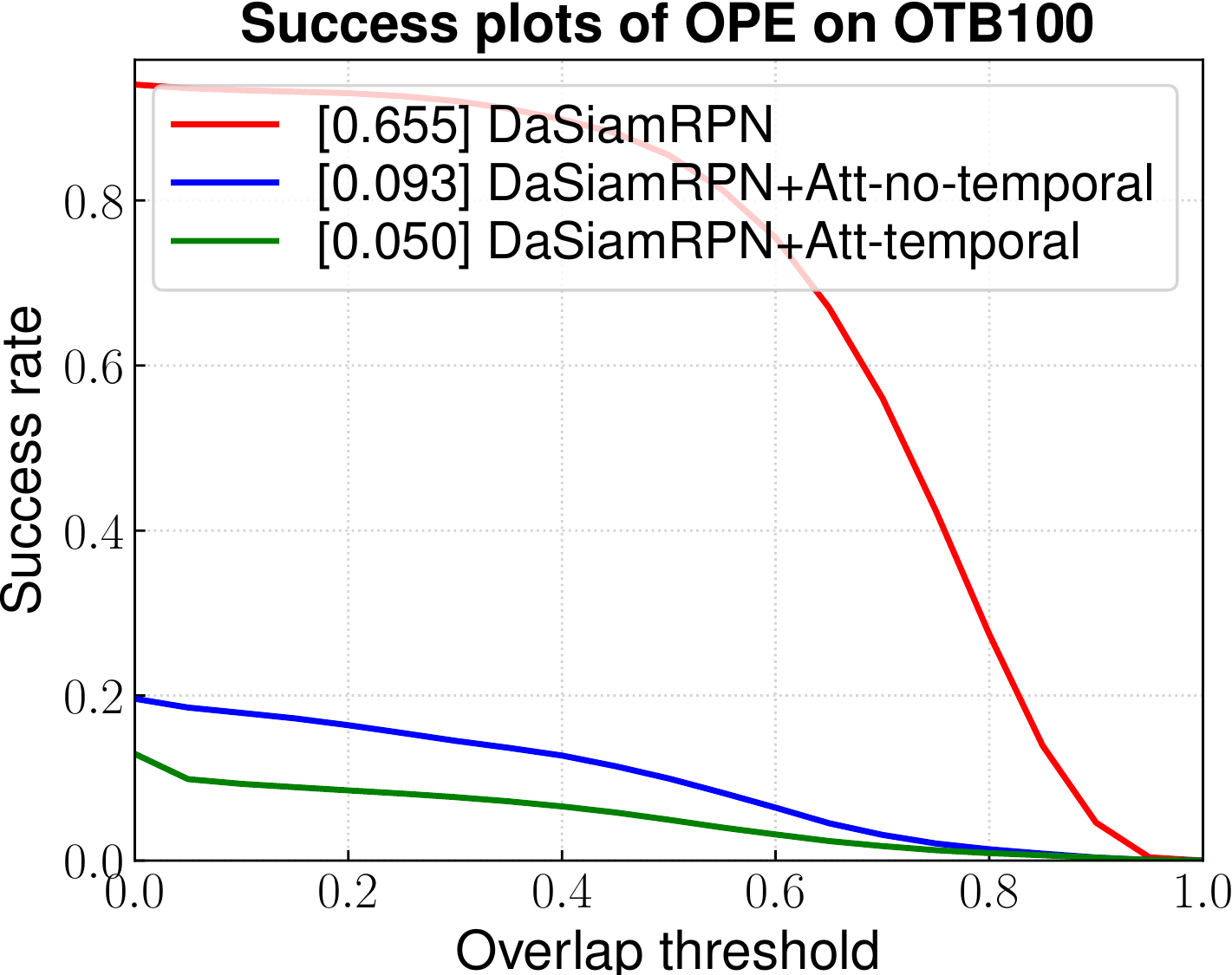}
		\end{tabular}
	\end{center}
	\caption{Temporal consistency validation of DaSiamRPN on the OTB100 dataset \cite{OTB-2015}.} 
	\label{fig:ab2}
\end{figure}

\begin{figure} [t]
	\centering 
	\subfigure[DaSiamRPN]
	{
		\includegraphics[width=\swfour]{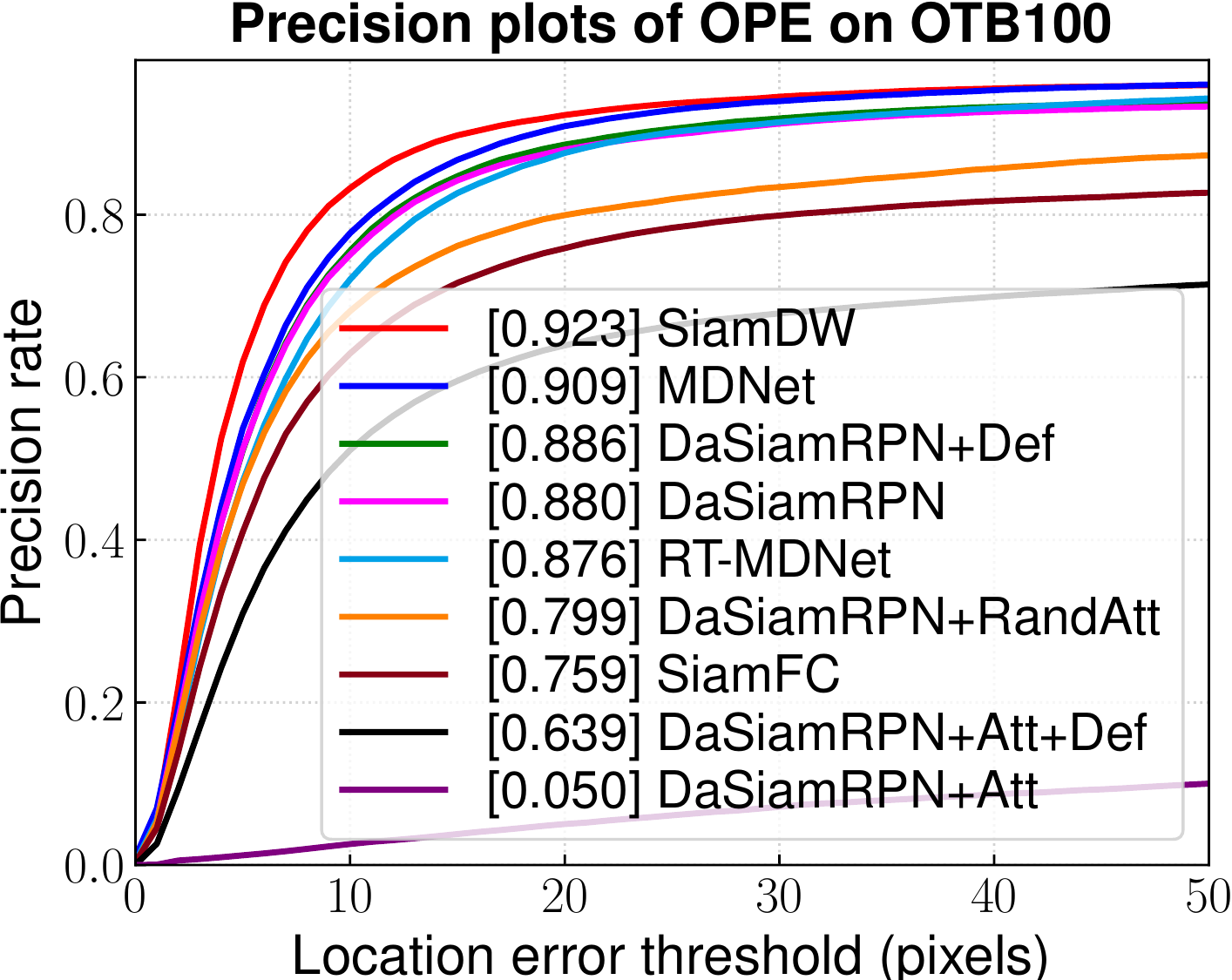}\ \ \ 
		\includegraphics[width=\swfour]{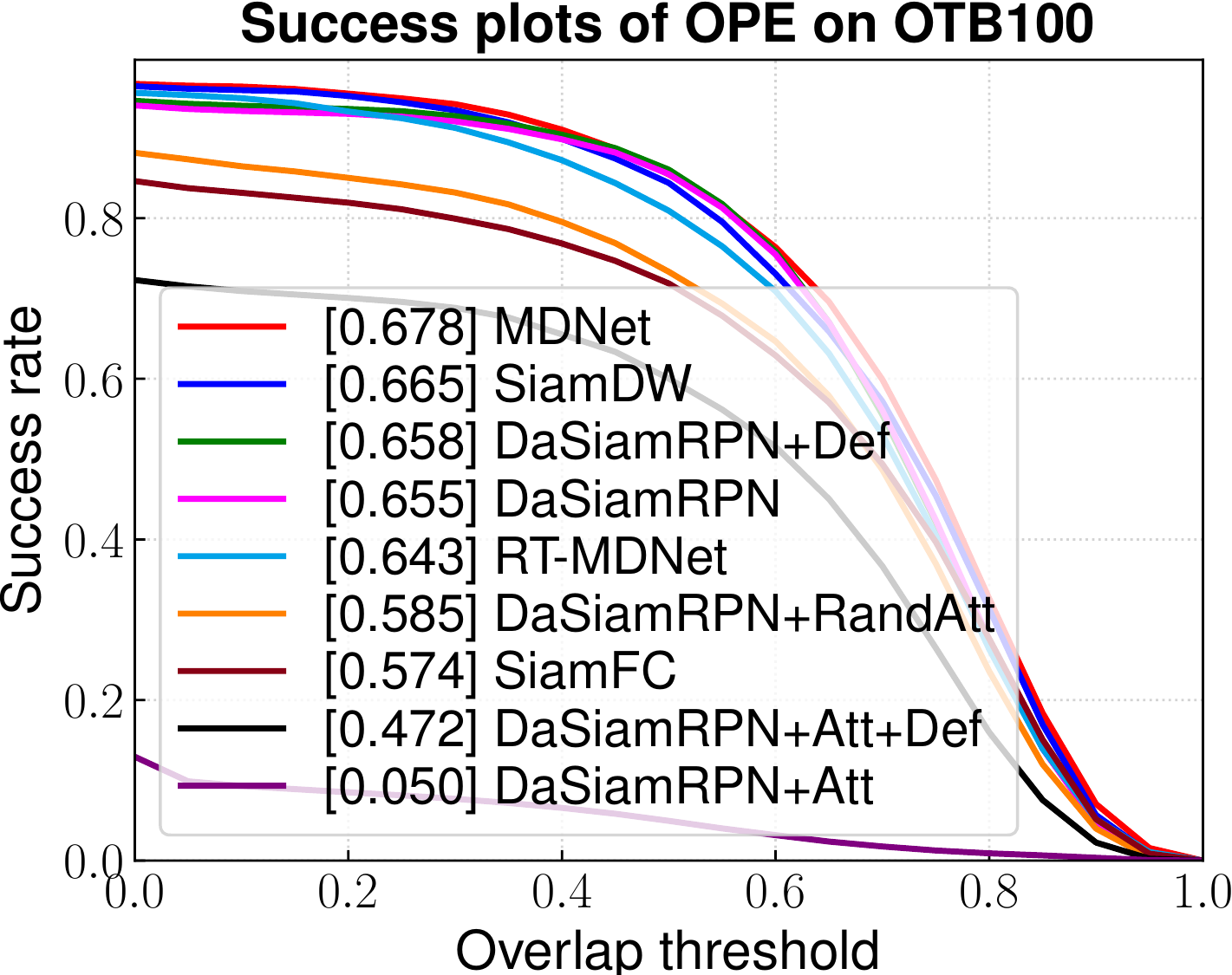}
	} 	
	\subfigure[RT-MDNet]{ 
		\includegraphics[width=\swfour]{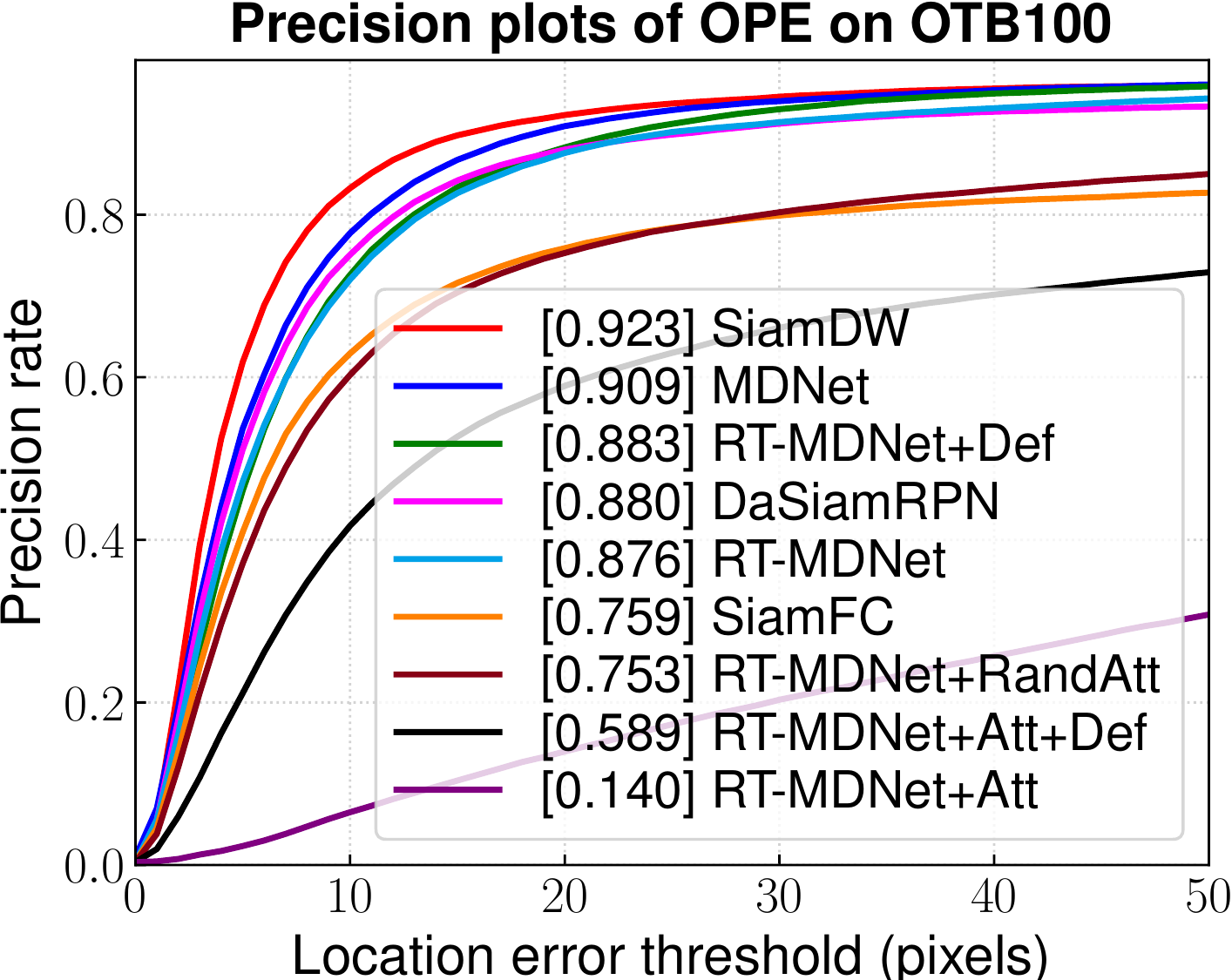}\ \ \ 
		\includegraphics[width=\swfour]{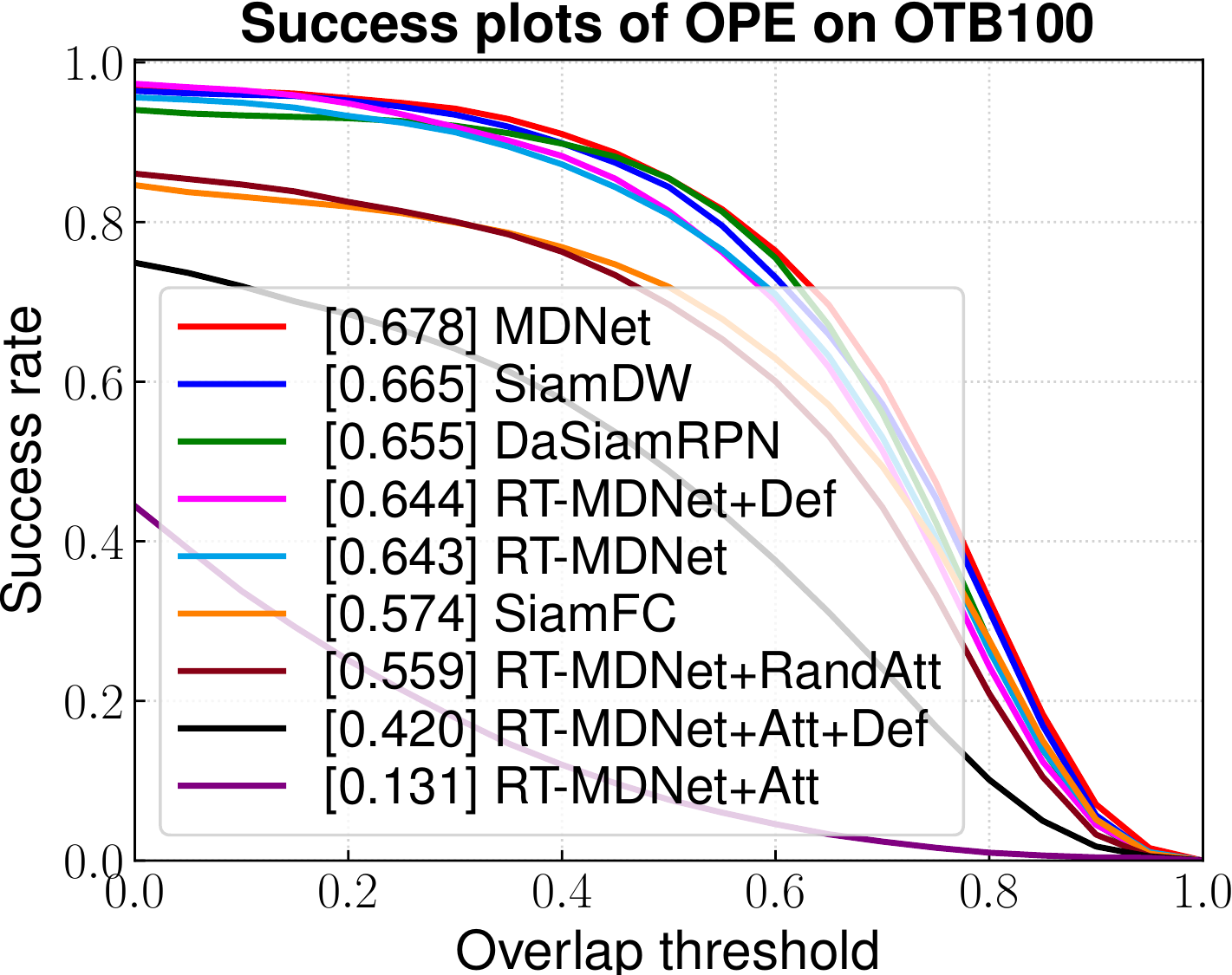}
	} 
	\caption{Tracking performance of the adversarial attack and defense methods on the OTB100 dataset. We denote Att and Def as the adversarial attack and defense, respectively, and denote Rand as random perturbations. 
	}
	\label{fig:plototb} 
\end{figure}
	
	\subsection{Ablation Studies}
	We analyze the effectiveness of each component of the proposed method. We take the DaSiamRPN tracker as a baseline as it contains both the classification and regression branches. We first evaluate the baseline performance on the OTB100 dataset. Then, we separately apply our attack method on the classification and regression branches. When we attack the regression branch, we analyze the offset and scale variation effects. Meanwhile, we combine both classification and regression attacks. Fig.~\ref{fig:ab1} shows the evaluation results where the attack on the regression branch degrades the performance more than the attack on the classification branch. Combining attacks on both branches leads to more degraded tracking performance.
	
	In addition to analyzing the attack in single frames, we show performance degradation via temporally consistent attack. We transfer the perturbation of the last frame to the current frame as initialization. Fig.~\ref{fig:ab2} shows that our temporally consistent attack decreases tracking accuracies more than the static image attack.

	\subsection{Benchmark Performance}
	
	We deploy the proposed attack and defense methods on DaSiamRPN and RT-MDNet on four standard benchmark datasets (OTB100~\cite{OTB-2015}, VOT-2016~\cite{vot2016}, VOT-2018~\cite{vot2018} and UAV123~\cite{uav123}). The results are presented as follows:
	
	{\flushleft\bf OTB100.} There are 100 video sequences in this dataset with substantial target variations. We apply the one-pass evaluation (OPE) with success and precision plots. The precision metric evaluates the ratio of frames whose center location error is within a certain threshold among all frames. The success plot measures the $IoU$ scores between the tracking results and ground truth at different thresholds. The area under the curve (AUC) success scores are also reported. 
	
	In order to evaluate our adversarial attack and defense methods, we first show the results of the baseline trackers. Second, we attack the baseline trackers by adding adversarial perturbations to input video sequences. In addition, we show the attack performance by adding random perturbations containing the same variations to those of adversarial perturbations. Fig.~\ref{fig:plototb} shows that our attack method reduces the AUC scores of DaSiamRPN from 0.655 to 0.050, and the AUC scores of RT-MDnet from 0.643 to 0.131. The baseline trackers under adversarial attacks perform much worse than that under random perturbations. This indicates the effectiveness of our adversarial attack method on baseline trackers. On the other hand, our defense method is able to suppress the maximal value of adversarial perturbations to restore the tracking performance. It recovers the AUC score of DaSiamRPN to 0.721 and RT-MDNet to 0.653. This demonstrates that our defense method can effectively remove the perturbations to restore tracking performance. In addition, we apply our defense method on original tracking sequences and improves AUC score from 0.880 to 0.886 for DaSiamRPN and from 0.876 to 0.883 for RT-MDNet. This indicates that perturbations (e.g., noise) exist in real world scenarios during image formation process (e.g., camera sensor noise, transformation from optical perception to digital storage). Our defense method is effective to estimate these naturally existing perturbations and eliminate their effects. 

	{\flushleft\bf VOT-2018.} There are 60 sequences on the VOT-2018 dataset. The VOT toolkit will reinitialize the tracker if it loses the target object during 5 consecutive frames. The evaluation metrics of VOT are expected average overlap (EAO), accuracy (Acc) and robustness (Rob). The accuracy represents the average overlap ratio and the robustness is measured by the number of reinitialization. EAO measures the overall performance of trackers.
	
	\begin{table}[t]
		\begin{center}
			\caption{Attack and defense on DaSiamRPN on VOT-2018 and VOT-2016 datasets.  } \label{table:table1}	
				\begin{tabular*} {1\linewidth} {p{0.33\linewidth}|p{0.1\linewidth}<{\centering}p{0.1\linewidth}<{\centering}p{0.11\linewidth}<{\centering}|p{0.1\linewidth}<{\centering}p{0.1\linewidth}<{\centering}p{0.11\linewidth}<{\centering}}
					\toprule
					\multirow{2}{*}{}&\multicolumn{3}{c|}{VOT-2018}&\multicolumn{3}{c}{VOT-2016}\\
					\cmidrule{2-7} 
					& Acc $\uparrow$ & Rob $\downarrow$& EAO $\uparrow$ & Acc  $\uparrow$ & Rob $\downarrow$& EAO $\uparrow$  \\
					\midrule
					\ DaSiamRPN &0.585 &0.272&0.380 &0.625 &0.224&0.439\\
					\ DaSiamRPN+RandAtt  &0.571&0.529 &0.223 &0.606&0.303 &0.336\\
					\ DaSiamRPN+Att   &0.536 & 1.447 & 0.097 &0.521 &1.613 & 0.078\\
					\ DaSiamRPN+Att+Def  &0.579 &0.674 &0.195&0.581 &0.722 &0.211\\
					\ \textbf{DaSiamRPN+Def } &0.584 &\textbf{0.253}  &\textbf{0.384} &0.622 &\textbf{0.214} &0.418 \\
					\bottomrule
				\end{tabular*}
		\end{center}
	\end{table}
	
	\begin{table}[t]
		\begin{center}
			\caption{Attack and defense on RT-MDNet on VOT-2018 and VOT-2016 datasets.  } \label{table:table2}	
				\begin{tabular*} {1\linewidth} {p{0.33\linewidth}|p{0.1\linewidth}<{\centering}p{0.1\linewidth}<{\centering}p{0.11\linewidth}<{\centering}|p{0.1\linewidth}<{\centering}p{0.1\linewidth}<{\centering}p{0.11\linewidth}<{\centering}}
					\toprule
					&\multicolumn{3}{c|}{VOT-2018}&\multicolumn{3}{c}{VOT-2016}\\
					\cmidrule{2-7}
					& Acc $\uparrow$ & Rob $\downarrow$& EAO $\uparrow$ & Acc  $\uparrow$ & Rob $\downarrow$& EAO $\uparrow$  \\
					\midrule
					\ RT-MDNet&0.533 &0.567&0.176&0.567 &0.196&0.370\\
					\ RT-MDNet+RandAtt&0.503&0.871 &0.137 &0.550&0.452 &0.235\\
					\ RT-MDNet+Att&0.475 &1.611 & 0.076&0.469 &0.928 & 0.128\\
					\ RT-MDNet+Att+Def &0.515 &1.021 &0.110 &0.531 &0.494  &0.225\\
					\ \textbf{RT-MDNet+Def}  &0.529 &\textbf{0.538}  &\textbf{0.179} &0.540 &\textbf{0.168}  &\textbf{0.374}\\
					\bottomrule
				\end{tabular*}
		\end{center}
	\end{table}

	Table~\ref{table:table1} and Table~\ref{table:table2} show the experimental results of DaSiamRPN and RT-MDnet. The failure number increases rapidly after attacking the original sequences. The EAO drops dramatically from 0.380 to 0.097 (i.e., a 74.5\% decrease) for DaSiamRPN and from 0.176 to 0.076 (i.e., a 56.8\% decrease) for RT-MDNet. By integrating our adversarial defense method, the EAO scores of DaSiamRPN and RT-MDNet are restored to 0.195 and 0.110, respectively. The performance decrease and restoration indicate the effectiveness of our adversarial attack and defense method. In addition, our defense method also improves the performance slightly when the baseline trackers are not under adversarial attacks. 
	
	{\flushleft\bf VOT-2016.} The VOT-2016 dataset consists of 60 sequences. The evaluation metrics are the same as those used in the VOT-2018 dataset.  Table~\ref{table:table1} and Table~\ref{table:table2} illustrate the performance of our attack and defense methods on DaSiamRPN and RT-MDNet. Our adversarial attack algorithm reduces the EAO scores by 82.2\% for DaSiamRPN and 65.4\% for RT-MDNet. With the use of the defense method on the adversarial examples, the accuracy of EAO is restored by 48.1\% for DaSiamRPN and 60.8\% for RT-MDNet. It indicates the effectiveness of both adversarial attack and defense methods. In addition, when applying our defense method to the attack-free deep trackers, the robustness gets largely improved. Due to the reinitialization scheme in VOT, we observe that EAO and robustness values decrease dramatically during attacks but accuracy value does not vary much. 
	
\begin{figure} [t]
		\centering 
		\subfigure[DaSiamRPN]
		{
			\includegraphics[width=\swfour]{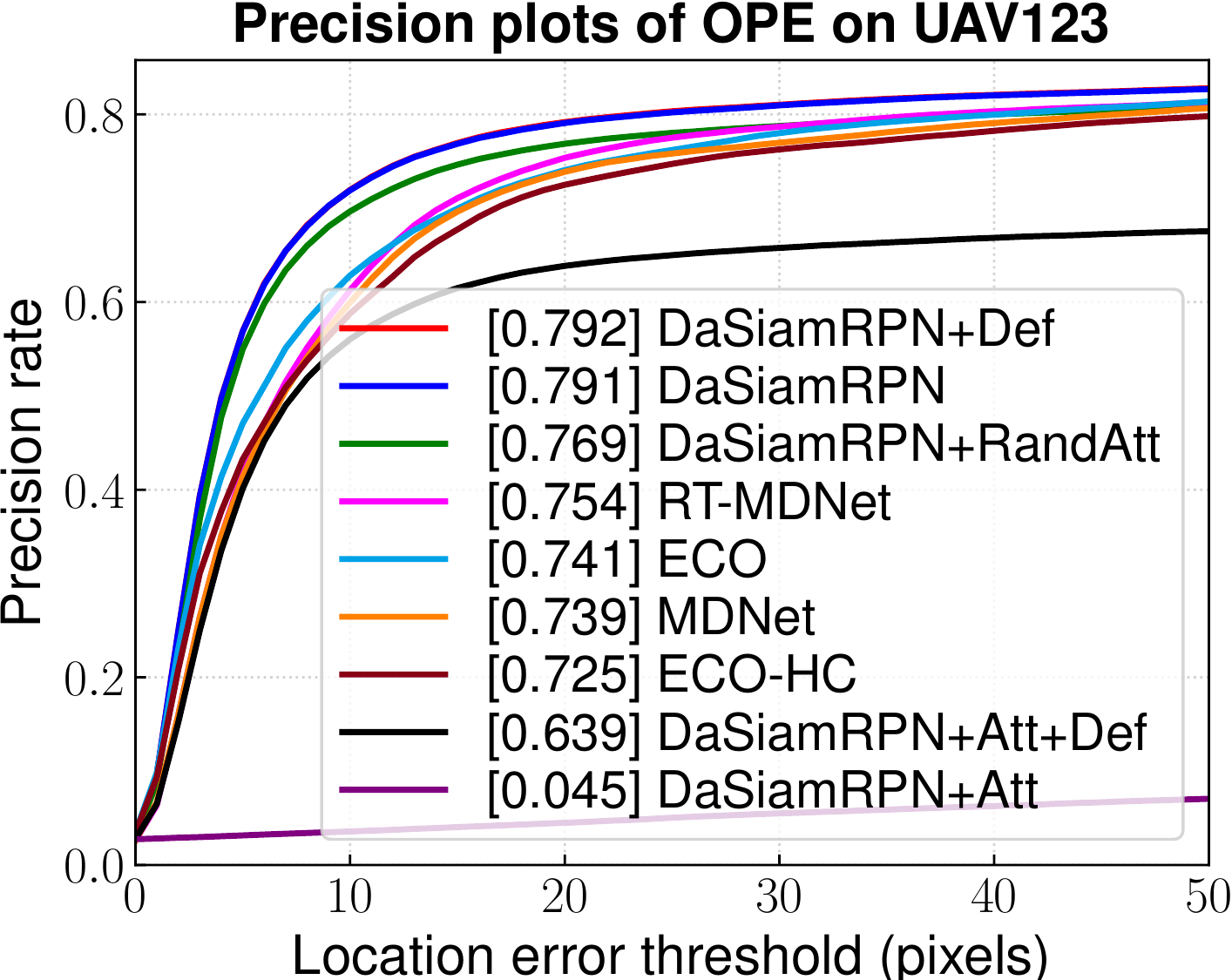}\ \ \ 
			\includegraphics[width=\swfour]{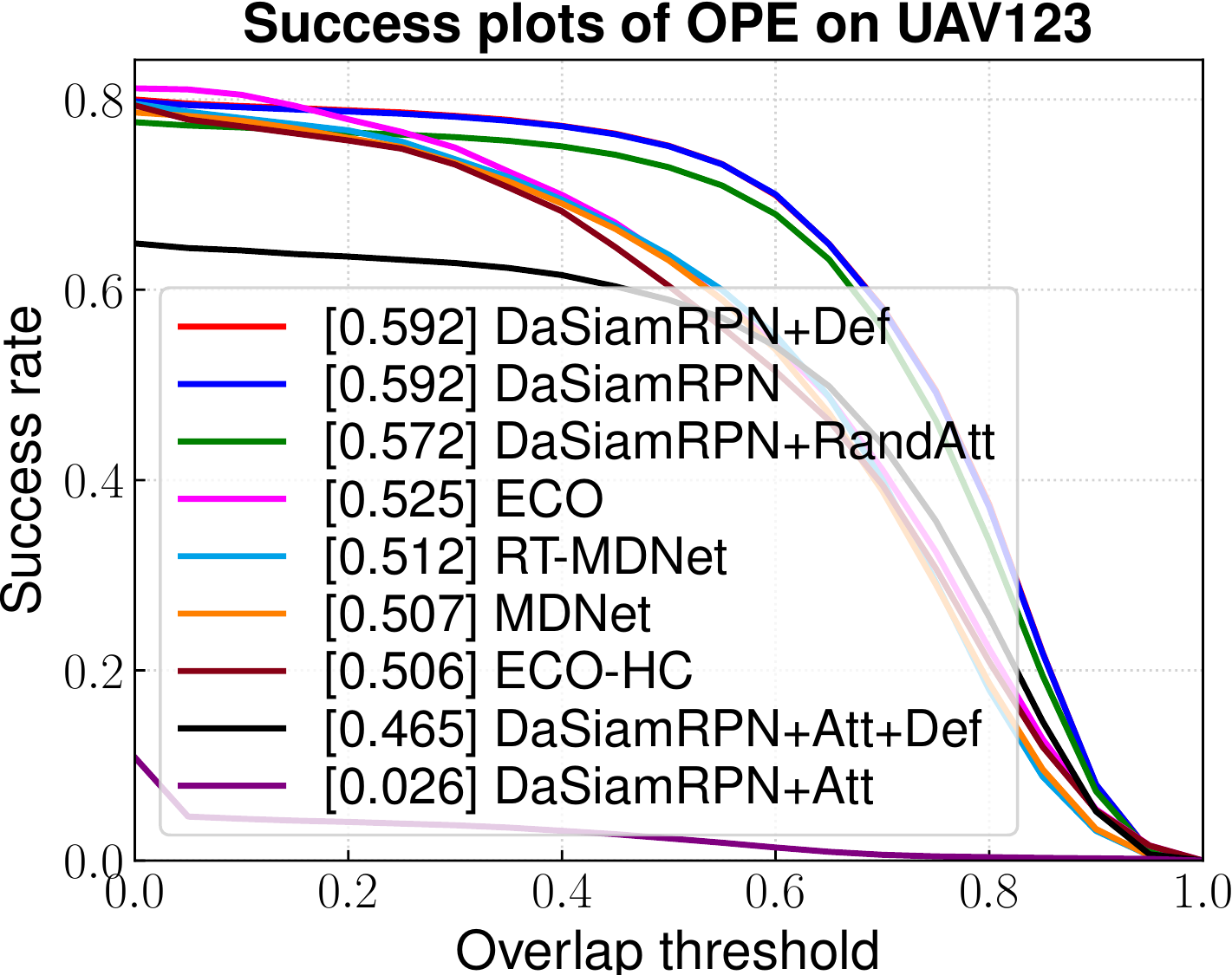}
		} 	
		
		\subfigure[RT-MDNet]{ 
			\includegraphics[width=\swfour]{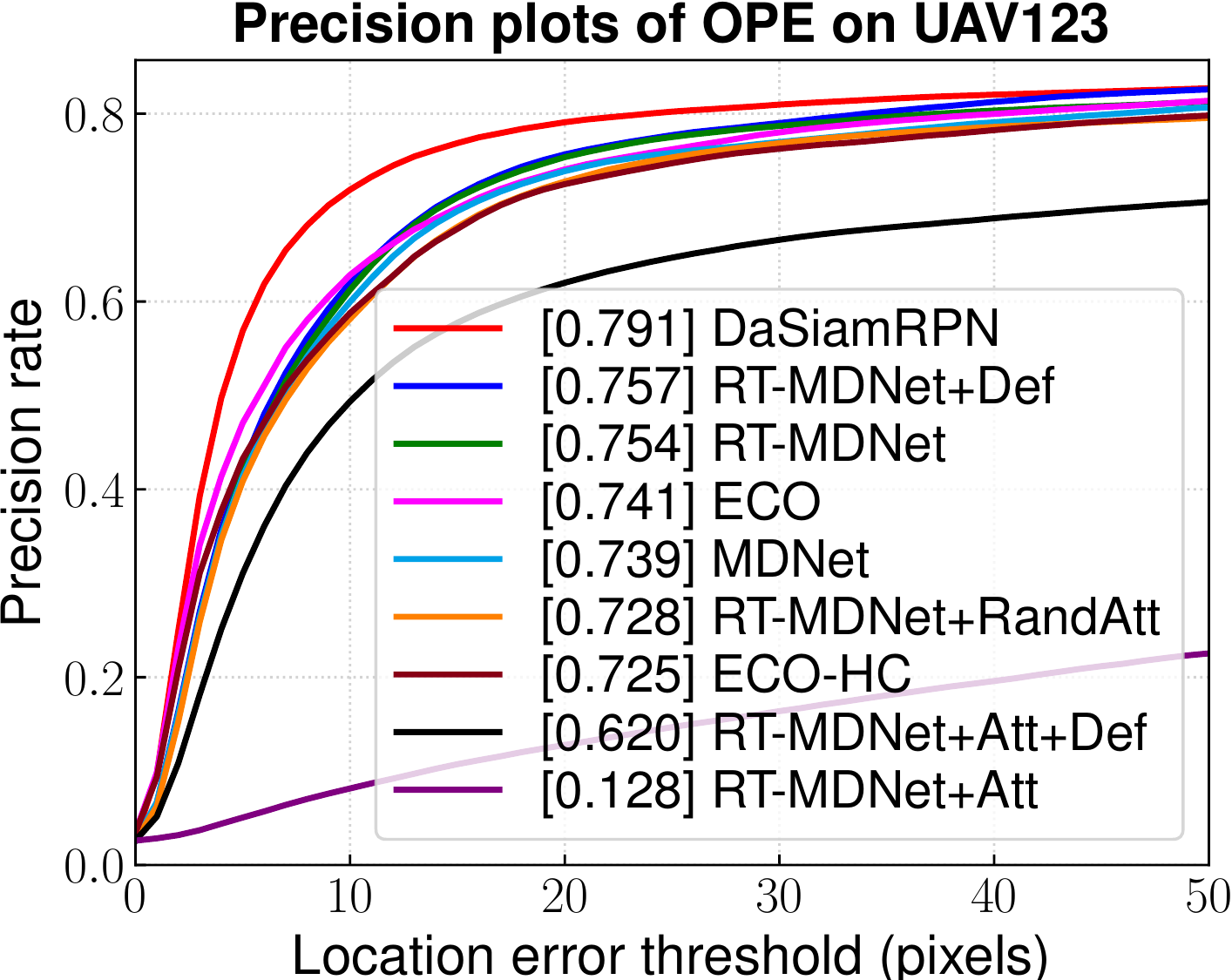}\ \ \ 
			\includegraphics[width=\swfour]{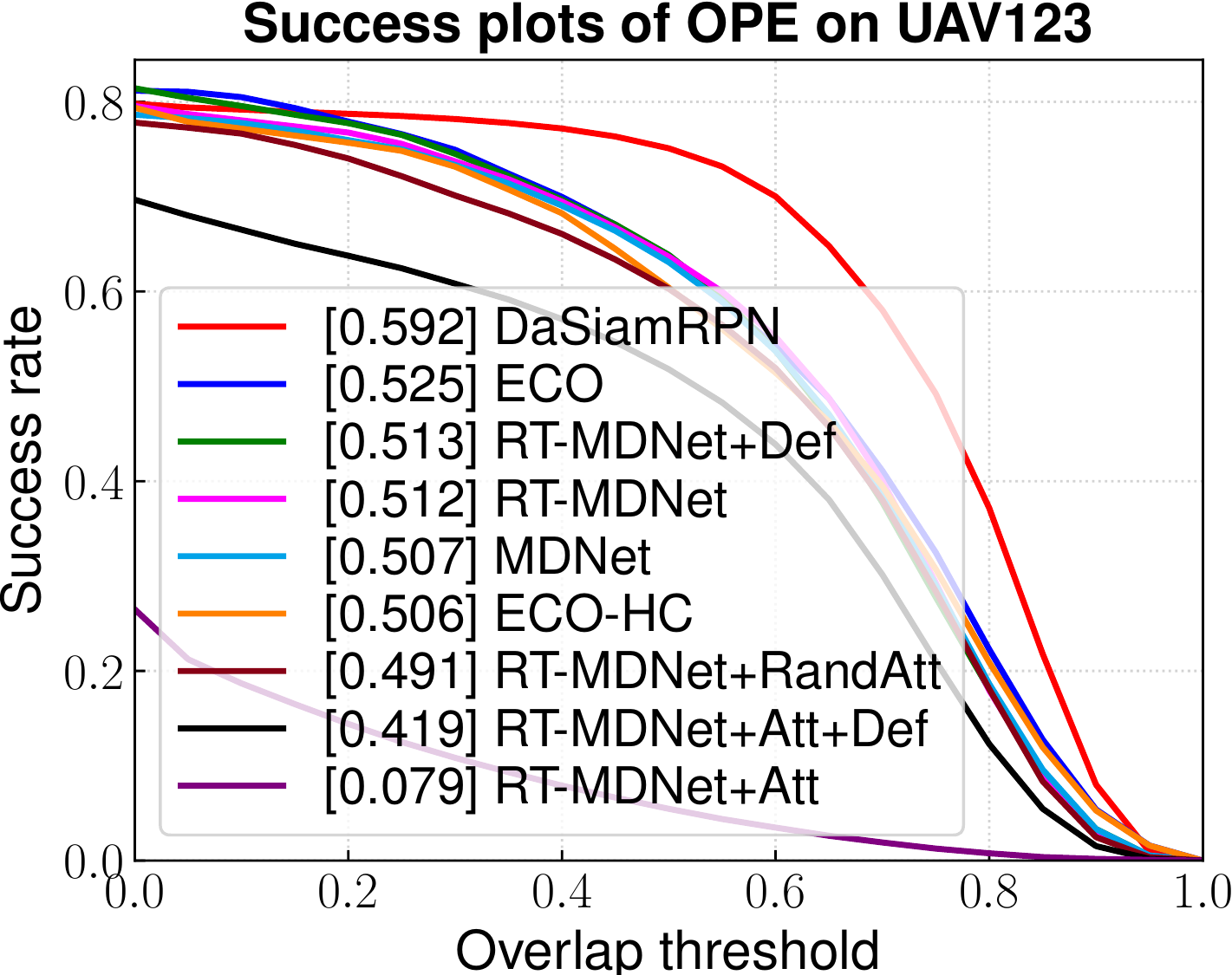}
		} 
		\caption{Results of the adversarial attack and defense methods on UAV123 dataset.}
		\label{fig:plotuav} 
	\end{figure}

	{\flushleft\bf UAV123.} The UAV123 dataset contains 123 sequences with more than 110K frames, which are captured from low-altitude unmanned aerial vehicles. We adopt the success and precision plots to evaluate the performance. Fig.~\ref{fig:plotuav} illustrates the precision and success plots of DaSiamRPN and RT-MDNet. Under attacks, the AUC scores drop by 95.6\% for DaSiamRPN and 84.6\% for RT-MDNet. The precision rate at 20 pixels is reduced by 94.3\% and 83.0\% for DaSiamRPN and RT-MDNet. The AUC scores are restored by 80.8\% and 82.2\% respectively after defending the adversarial examples since the target objects on the UAV123 dataset mostly undergo large shape changes. 

	\def\swthree{0.19\linewidth}
	\begin{figure*}[t]
		\begin{center}
			\includegraphics[width=0.40\linewidth]{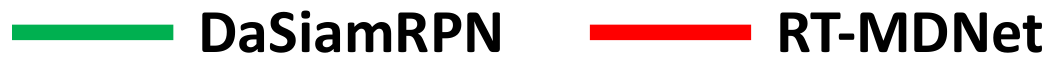}\\
		\end{center}
		\begin{center}
		\begin{tabular}{ccccc}
			\includegraphics[width=\swthree]{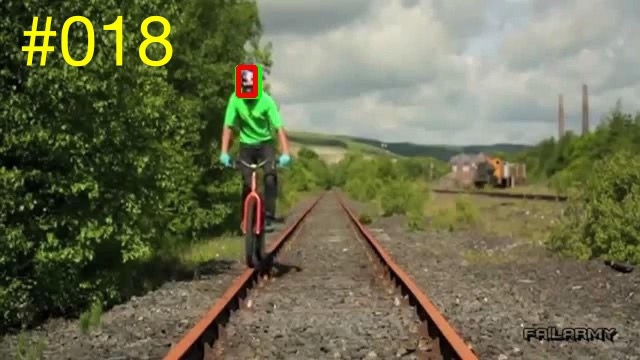}&
			\includegraphics[width=\swthree]{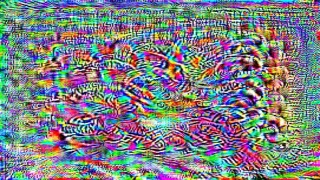}&
			\includegraphics[width=\swthree]{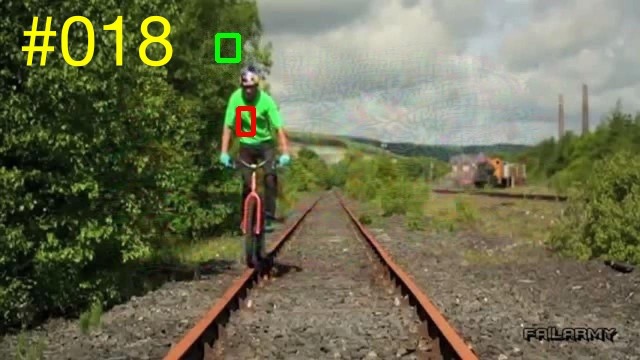}&
			\includegraphics[width=\swthree]{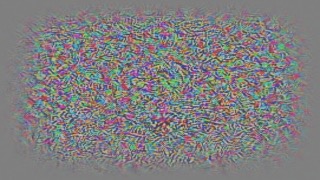}&
			\includegraphics[width=\swthree]{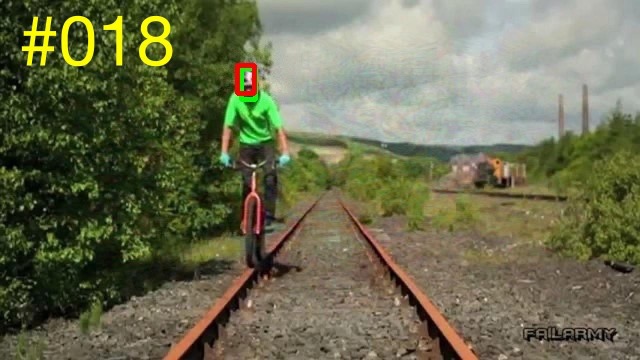}\\
			\includegraphics[width=\swthree]{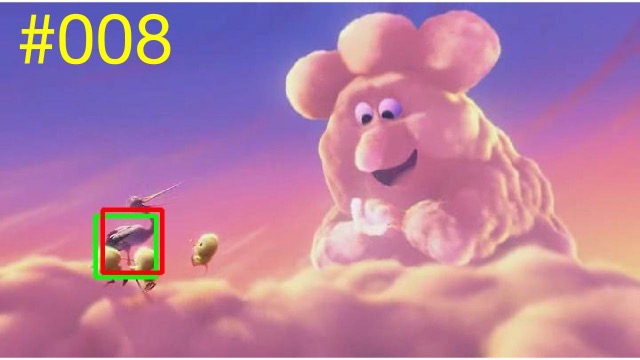}&
			\includegraphics[width=\swthree]{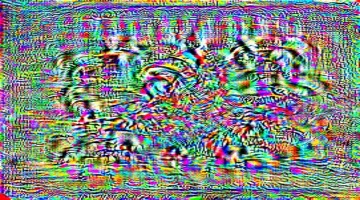}&
			\includegraphics[width=\swthree]{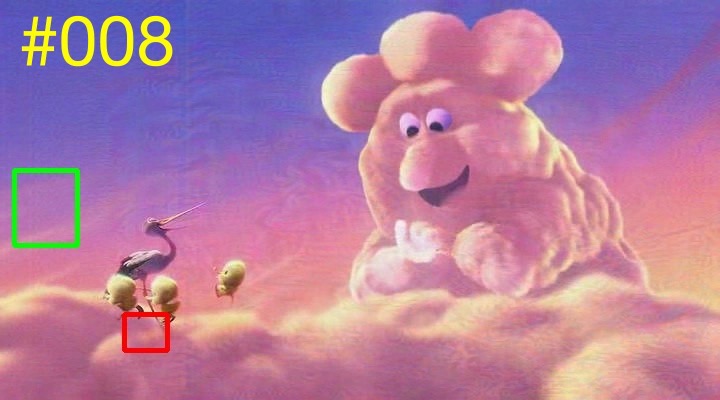}&
			\includegraphics[width=\swthree]{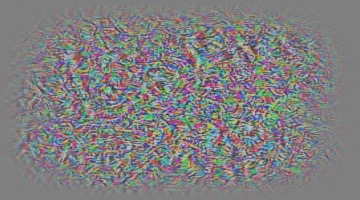}&
			\includegraphics[width=\swthree]{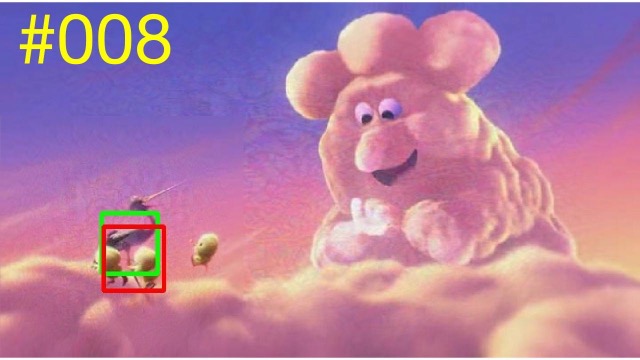}\\
			\includegraphics[width=\swthree]{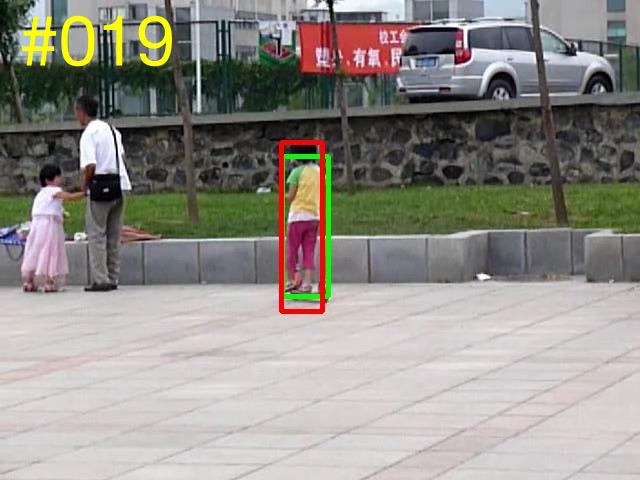}&
			\includegraphics[width=\swthree]{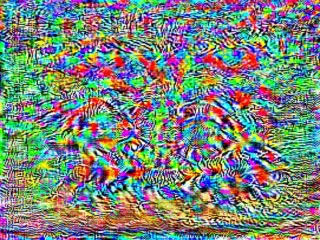}&
			\includegraphics[width=\swthree]{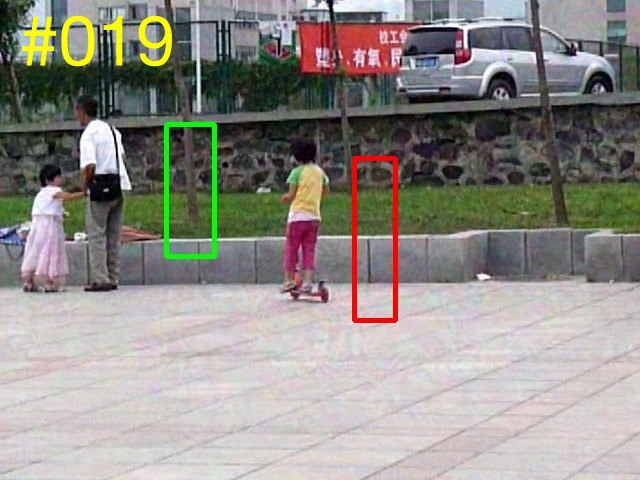}&
			\includegraphics[width=\swthree]{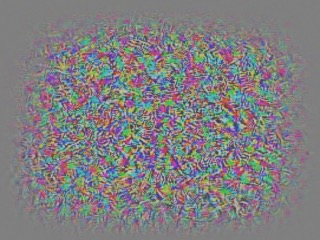}&
			\includegraphics[width=\swthree]{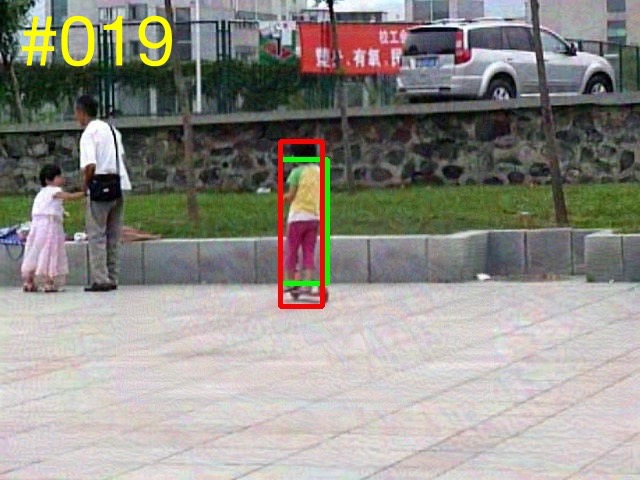}\\
			\includegraphics[width=\swthree]{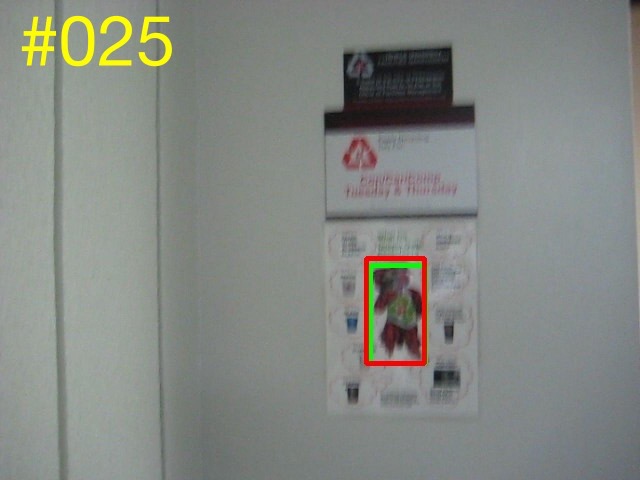}&
			\includegraphics[width=\swthree]{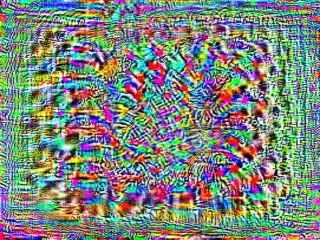}&
			\includegraphics[width=\swthree]{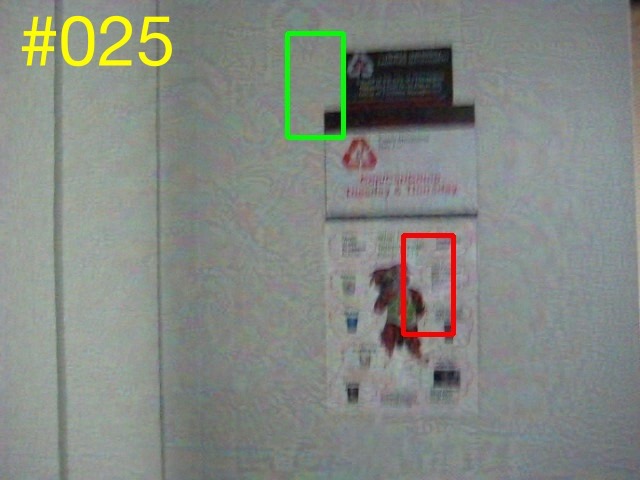}&
			\includegraphics[width=\swthree]{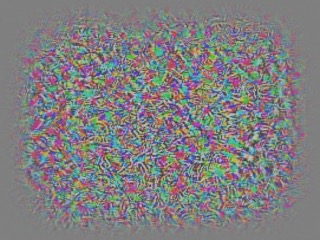}&
			\includegraphics[width=\swthree]{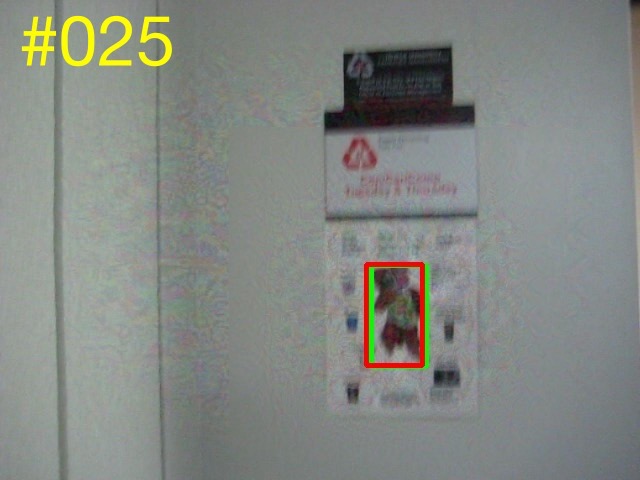}\\
			\includegraphics[width=\swthree]{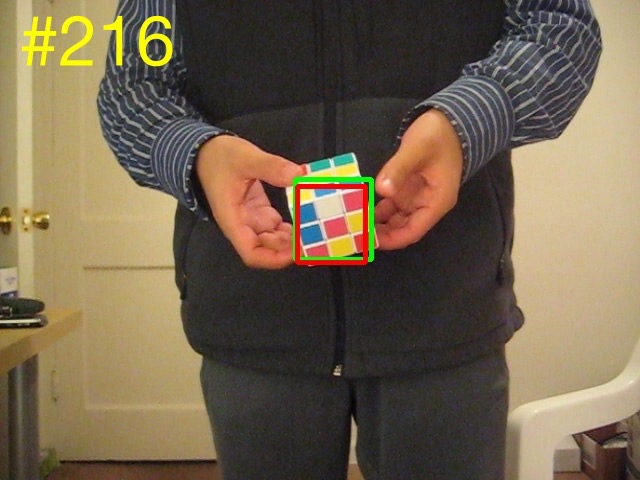}&
			\includegraphics[width=\swthree]{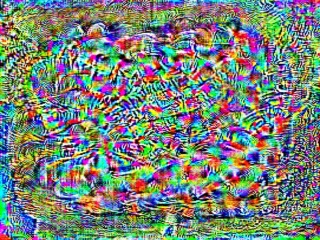}&
			\includegraphics[width=\swthree]{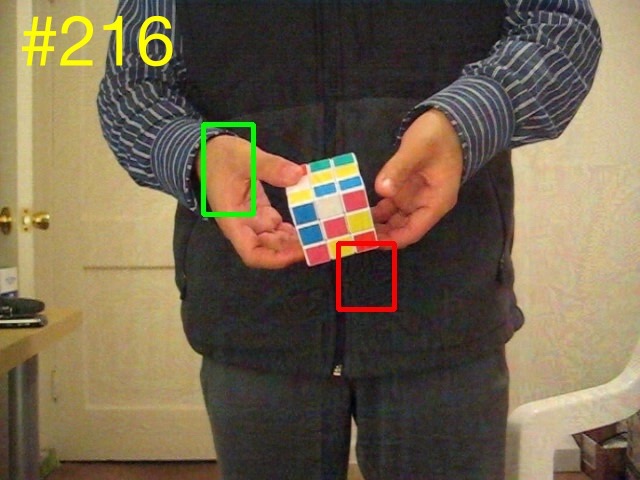}&
			\includegraphics[width=\swthree]{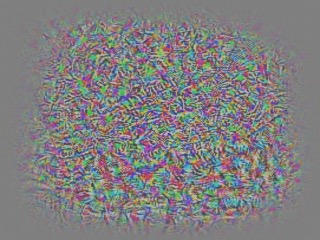}&
			\includegraphics[width=\swthree]{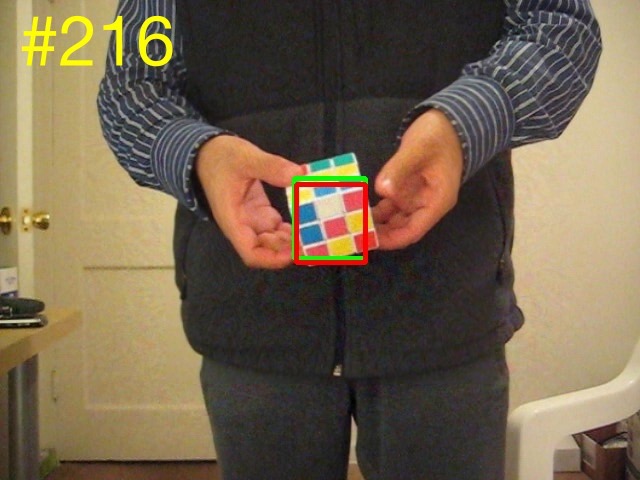}\\
			(a) Input&(b) Adversarial&(c) Attack&(d) Estimated &(e) Defense\\
			frames& perturbations& results& perturbations& results\\
		\end{tabular}
		\end{center}
		\caption{Qualitative evaluation of adversarial attack and defense on video sequences.}
		\label{fig:10}
	\end{figure*}
	 
	\subsection{Qualitative Evaluations}
	Fig.~\ref{fig:10} qualitatively shows the tracking results of our adversarial attack and defense methods for DaSiamPRN~\cite{zhu-eccv18-dasiamrpn} and RT-MDNet~\cite{jung-eccv18-rtmdnet} on 5 challenging sequences. We visualize the attack and defense perturbations in (b) and (d). 
	In the original sequences shown in (a), both DaSiamRPN and RT-MDNet locate the target objects and estimate the scale changes accurately. After injecting adversarial perturbations in (b), DaSiamRPN fails to track the target and RT-MDNet estimates the target scale inaccurately as shown in (c). The tracking accuracy of RT-MDNet does not degrade severely compared to DaSiamRPN because RT-MDNet only contains the classification branch while DaSiamRPN contains both classification and regression branches. When we perform defense, the adversarial perturbations are estimated and shown in (d). By subtracting the estimated perturbations from the adversarial examples, DaSiamRPN and RT-MDNet are able to locate target objects correctly in the video sequences shown in (e).
 
	\section{Concluding Remarks}
	In this paper, we propose the adversarial attack and defense methods by generating lightweight perturbations within the deep tracking framework. When generating adversarial examples, we integrate the temporal perturbations into frames by perplexing trackers with indistinguishable correct and incorrect inferences. When defending adversarial examples, we suppress the maximum of adversarial perturbation to restore the tracking accuracy. Extensive experiments on four standard benchmarks demonstrate that the proposed methods perform favorably both in adversarial attack and defense. In addition, our defense method is capable of reducing interference from perturbations in the real world scenarios to robustify deep trackers.
	
	\subsubsection{Acknowledgements.}
	This work was supported by National Key Research and Development Program of China (2016YFB1001003), NSFC (U19B2035, 61527804, 60906119), STCSM (18DZ1112300). C. Ma was sponsored by Shanghai Pujiang Program.

	\clearpage
	%
	%
	
	\bibliographystyle{splncs04}
	\bibliography{egbib}
	
\end{document}